\documentclass[runningheads]{llncs}
\usepackage{scrextend}
\usepackage[super]{nth}
\usepackage{caption}
\usepackage{subcaption}
\usepackage{graphicx}
\usepackage{color}
\usepackage{amsmath,amssymb,amsfonts}
\usepackage[inline]{enumitem}
\usepackage{csquotes}
\usepackage{wrapfig,lipsum,booktabs}
\usepackage{placeins}
\usepackage{hyperref}

\newcommand{\pt}{T{=}(V,\allowbreak E,\allowbreak \lambda,\allowbreak r)}

\newcommand{\emptyseq}{\ensuremath{\langle\rangle}}

\usepackage[ruled,lined,vlined]{algorithm2e}

\SetCommentSty{xCommentSty}

\usepackage{tikz}
\usetikzlibrary{fit}
\usetikzlibrary{arrows.meta}
\usepackage{eqparbox}
\newbox\nodebox
\usetikzlibrary{trees,shapes.geometric,calc}
\usetikzlibrary{positioning}
\pgfdeclarelayer{back}
\pgfsetlayers{back,main}

\usepackage{colortbl}
\definecolor{logMove}{HTML}{000000}
\definecolor{modelMove}{HTML}{666666}
\definecolor{synchronousMove}{HTML}{d9d9d9}
\usepackage{xcolor}
\colorlet{comment}{blue}
\newcommand{\cfbox}[2]{%
    \colorlet{currentcolor}{.}%
    {\color{#1}%
    \fbox{\color{currentcolor}#2}}%
}

\usepackage{makecell}
\usepackage{tabularx}
\usepackage{multicol}
\begin{document}
\title{Freezing Sub-Models During Incremental Process Discovery: Extended Version\thanks{This paper is an extended version of the paper \emph{Freezing Sub-Models During Incremental Process Discovery} presented at the 40th International Conference on Conceptual Modeling 2021}}
%
%
\author{Daniel Schuster\inst{1,2} \orcidID{0000-0002-6512-9580} \and
Sebastiaan J. van Zelst\inst{1,2} \orcidID{0000-0003-0415-1036} \and
Wil M. P. van der Aalst\inst{1,2} \orcidID{0000-0002-0955-6940} }

\authorrunning{D. Schuster et al.}
%
\institute{Fraunhofer Institute for Applied Information Technology FIT, Germany\\
\email{\{daniel.schuster,sebastiaan.van.zelst,wil.van.der.aalst\}@fit.fraunhofer.de}
\and
RWTH Aachen University, Aachen, Germany\\
}
\maketitle              
\begin{abstract}

Process discovery aims to learn a process model from observed process behavior.
From a user's perspective, most discovery algorithms work like a \emph{black box}.
Besides parameter tuning, there is no interaction between the user and the algorithm.
Interactive process discovery allows the user to exploit domain knowledge and to guide the discovery process.
Previously, an incremental discovery approach has been introduced where a model, considered to be \enquote{under construction}, gets incrementally extended by user-selected process behavior.
This paper introduces a novel approach that additionally allows the user to \emph{freeze} model parts within the model under construction.
Frozen sub-models are not altered by the incremental approach when new behavior is added to the model.
The user can thus steer the discovery algorithm.
Our experiments show that freezing sub-models can lead to higher quality models.


\end{abstract}
\section{Introduction}

Executing business processes generates valuable data in the information systems of organizations. 
\emph{Process mining} comprises techniques to analyze these \emph{event data} and aims to extract insights into the executed processes to improve them~\cite{DBLP:books/sp/Aalst16}. 
This paper focuses on \emph{process discovery}, a key discipline within process mining.

Conventional process discovery algorithms use observed process behavior, i.e., event data, as input and return a process model that describes the process, as recorded by the event data. 
Since event data often have quality issues---for instance, incomplete behavior, noise, or wrongly recorded process behavior---process discovery is a challenging task. 
Apart from modifying the input (event data) or the subsequent alteration of the output (discovered model), the user has no options to interact with the algorithm.
Thus, conventional process discovery works like a black box from a user’s perspective.

To overcome this limitation, the field of \emph{interactive process discovery} has emerged. 
The central idea is to exploit the domain knowledge of process participants within process discovery in addition to the standard input of event data. 
Several techniques have been proposed. 
However, most approaches to date only attempt to use additional inputs besides the event data. 
Thus, a user still has only limited options to influence the algorithm during the actual discovery phase, and the discovery algorithm remains a black box from a user's perspective.

Recently, we have introduced an incremental process discovery framework allowing a user to incrementally add process behavior to a process model under construction~\cite{10.1007/978-3-030-50316-1_25}. 
This framework enables the user to control the algorithm any time by interactively deciding on the process behavior to be added next. 

\begin{figure}[t]
    \centering
    \includegraphics[width=\textwidth,trim=0cm 7.7cm 2.3cm 0cm,clip]{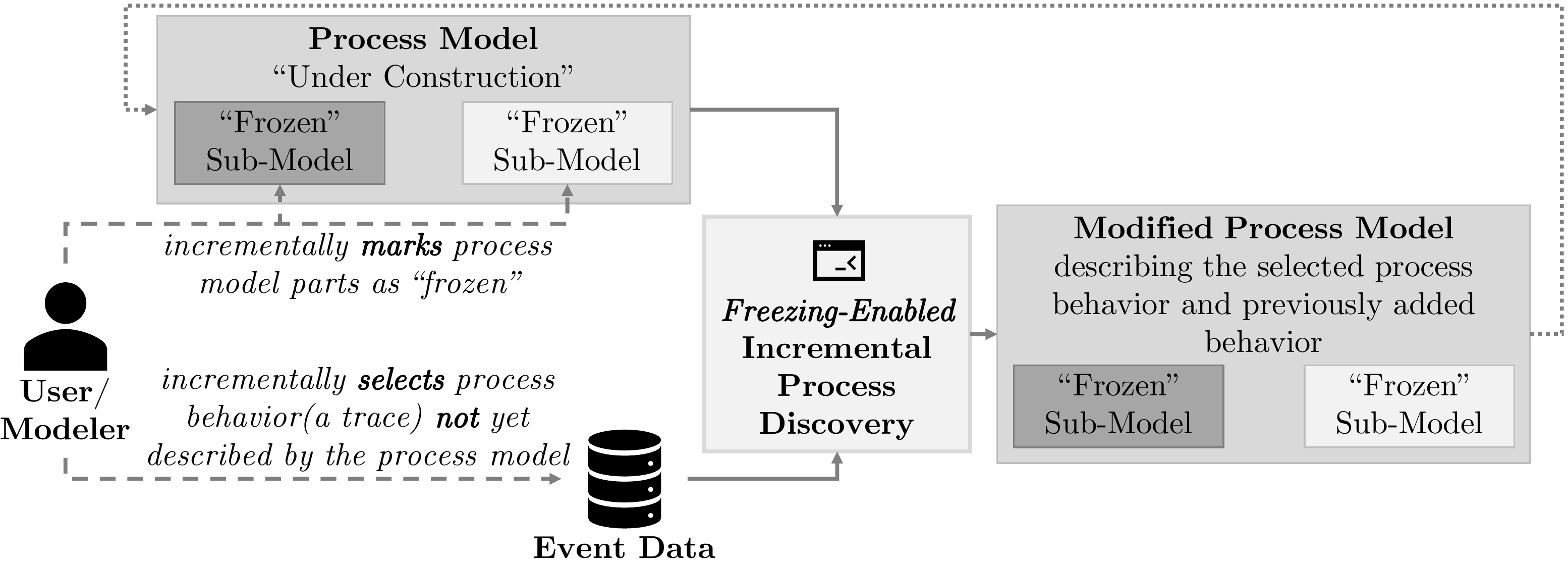}
    \caption{Overview of the proposed freezing option extending \emph{incremental} process discovery. A user incrementally selects traces from the event log and optionally freezes sub-models that should not get altered in the model \enquote{under construction}}
    \label{fig:freezing_overview}
\end{figure}

In the context of incremental process discovery, we propose in this paper a novel way to interact with a process discovery algorithm as a user.
During the discovery phase, we allow a user to freeze sub-models of the process model under construction.
By marking sub-models as frozen, the incremental process discovery approach does not alter these frozen model parts during the ongoing incremental discovery.
\autoref{fig:freezing_overview} summarizes the approach.
Our proposed approach can be applied with any incremental process discovery algorithm.

Many use cases exist where freezing sub-models is beneficial in the context of incremental process discovery. 
For instance, it enables a user to combine \emph{de jure} and \emph{de facto} process models \cite{DBLP:books/sp/Aalst16}. 
A de jure model describes how a process should be executed (normative), and a de facto model describes how a process was executed (descriptive). 
A user might freeze a process model part because, from the user’s perspective, the sub-model to be frozen is already normative, i.e., it already describes a certain process part as it should be executed. 
Therefore, a user wants to protect this sub-model from being further altered while incrementally adding new behavior to the process model under construction. 
Similarly, a user could start with predefined sub-models that are frozen, i.e., de jure models, and incrementally discover missing parts around the predefined ones. 
Thus, the proposed freezing option allows combining process discovery with process modeling.
Our conducted experiments show that freezing sub-models can lead to higher quality models.
This demonstrates that the freezing option is a novel and useful form of user-interaction in the area of interactive process discovery.

The remainder of this paper is structured as follows. 
\autoref{sec:related_work} presents related work while \autoref{sec:preliminaries} presents preliminaries. 
\autoref{sec:freezing} presents the proposed approach of freezing sub-models during incremental process discovery. 
\autoref{sec:evaluation} present an experimental evaluation, and \autoref{sec:conclusion} concludes this paper.

\section{Related Work}
\label{sec:related_work}

This section mainly focuses on interactive process discovery.
For an overview of process mining and conventional process discovery, we refer to~\cite{DBLP:books/sp/Aalst16}.

In~\cite{10.1145/2710020}, the authors propose to incorporate precedence constraints over the activities within process discovery. 
In \cite{DBLP:conf/simpda/DixitBAHB15a}, an approach is presented where an already existing process model is post-processed s.t. user-defined constraints are fulfilled.
In \cite{10.1007/978-3-642-45005-1_23}, an approach is presented where domain knowledge in form of a process model is given. 
From the initially given model, which reflects the domain knowledge, and the event data, a new model is discovered. 
Compared to our extended incremental process discovery, all approaches remain a black-box from a user's perspective since they work in a fully automated fashion. 
In~\cite{DBLP:conf/er/DixitVBA18}, an interactive modeling approach is proposed. 
A user constructs the model guided by the algorithm, i.e., the user makes the design decisions in the process model, as opposed to our approach, where the discovery algorithm is guided by the user.

Related work can also be found in the area of process repair~\cite{10.1007/978-3-642-32885-5_19}. 
However, the setting of process model repair, where the repaired model is tried to be as similar as possible to the original one, differs from incremental process discovery.
In~\cite{10.1007/978-3-319-69462-7_5} an interactive and incremental repair approach is proposed.
Deviations are visualized to a user for a process model and a given event log and the user has to manually repair the deviations under the guidance of the approach.


\section{Preliminaries}
\label{sec:preliminaries}

We denote the power set of a given set $X$ by $\mathcal{P}(X)$.
We denote the universe of multi-sets over a set $X$ by $\mathcal{B}(X)$ and the set of all sequences over $X$ as $X^*$, e.g., $\langle a,b,b \rangle{\in} \{a,b,c\}^*$. 
Given two sequences $\sigma$ and $\sigma'$, we denote their concatenation by $\sigma {\cdot} \sigma'$, e.g., $\langle a \rangle {\cdot} \langle b,c\rangle {=} \langle a,b,c\rangle $.
We extend the $\cdot$ operator to sets of sequences, i.e., let $S_1,S_2{\subseteq}X^*$ then $S_1{\cdot}S_2{=}\allowbreak\{\sigma_1{\cdot}\sigma_2\ {|} \sigma_1{\in}S_1 {\wedge}\allowbreak \sigma_2{\in}S_2\}$.
For sequences $\sigma,\sigma'$, the set of all interleaved sequences is denoted by $\sigma {\diamond} \sigma'$, e.g., $\langle a,b \rangle {\diamond} \langle c \rangle {=} \{ \langle a,b,c\rangle,\allowbreak\langle a,c,b\rangle,\allowbreak\langle c,a,b\rangle \}$.
We extend the $\diamond$ operator to sets of sequences. 
Let $S_1,S_2 {\subseteq} X^*$, $S_1{\diamond} S_2$ denotes the set of interleaved sequences, i.e., $S_1{\diamond} S_2 {=}\allowbreak {\bigcup}_{\sigma_1{\in}S_1,\sigma_2{\in}S_2} \sigma_1 {\diamond} \sigma_2$.

For $\sigma{\in}X^*$ and $X'{\subseteq}{X}$, we define the projection function $\sigma_{\downarrow_{X'}} {:} X^* {\to}\allowbreak (X')^*$ with: $\emptyseq_{\downarrow_{X'}} {=} \emptyseq$,\allowbreak $\big(\langle x\rangle {\cdot} \sigma\big)_{\downarrow_{X'}} {=} \langle x\rangle {\cdot} \sigma_{\downarrow_{X'}}$ if $x{\in}X'$ and $(\langle x\rangle {\cdot} \sigma)_{\downarrow_{X'}} {=} \sigma_{\downarrow_{X'}}$ otherwise.

Let $t{=}(x_1,\dots,x_n){\in}X_1{\times}  \dots {\times} X_n$ be an $n$-tuple over $n$ sets.
We define projection functions that extract a specific element of $t$, i.e., $\pi_1(t){=}x_1,\dots, \pi_n(t){=}x_n$, e.g., $\pi_2\left(\left( a,b,c \right)\right){=}b$. 
Analogously, given a sequence of length $m$ with $n$-tuples $\sigma{=}\langle(x^1_1,\dots,x^1_n),\dots, \allowbreak (x_1^m,\dots,x_n^m)\rangle$, we define $\pi^*_1(\sigma){=}\langle x_1^1,\dots,\allowbreak x_1^m\rangle, \dots,\allowbreak\pi^*_n(\sigma){=}\allowbreak\langle x_n^1,\dots,x_n^m\rangle$.
For instance,  $\pi^*_2\big(\langle (a,b), (a,c), (b,a) \rangle\big) {=} \langle b,c,a \rangle$.

\subsection{Event Data and Process Models}

\begin{table}[tb]
\scriptsize
\caption{Example of an event log from an e-commerce process}
\label{tab:event_log}
\centering
\begin{tabular}{|c|c|c|c|}
\hline
\textbf{Case-ID} & \textbf{Activity}    & \textbf{Timestamp} & \textbf{$\cdots$} \\ \hline
151              & place order (p)     & 10/03/21 12:00     & $\cdots$          \\ \hline
153              & cancel order (c)     & 10/03/21 12:24     & $\cdots$          \\ \hline
152              & place order (p)     & 11/03/21 09:11     & $\cdots$          \\ \hline
151              & payment received (r) & 11/03/21 10:00     & $\cdots$          \\ \hline
$\cdots$         & $\cdots$             & $\cdots$           & $\cdots$          \\ \hline
\end{tabular}
\end{table}

The data that are generated during the execution of (business) processes and stored in information systems are called \emph{event data}~\cite{DBLP:books/sp/Aalst16}. 
\autoref{tab:event_log} shows an example of an event log.
Each row represents an event.
Events with the same case-id belong to the same process execution often referred to as a \emph{case}.
The sequence of executed activities for a case is referred to as a \emph{trace}, e.g., the partial trace for case 151 is: $\langle p, r,\dots \rangle$. 
Next, we formally define an event log as a multi-set of traces.
Note that the same trace can occur multiple times in an event log. 
\begin{definition}[Event Log]
Let $\mathcal{A}$ be the universe of activities. 
$L{\in}\mathcal{B}(\mathcal{A}^*)$ is an event log. 
\end{definition}

Process models allow us to specify the control flow of a process.
In this paper, we use process trees~\cite{DBLP:books/sp/Aalst16}, e.g., see \autoref{fig:process_tree_example}. 
Leaves represent activities and $\tau$ represents an unobservable activity, needed for certain control flow patterns.
Inner nodes represent operators that specify the control flow among their subtrees.
Four operators exist: \emph{sequence} ($\rightarrow$), \emph{excl. choice} ($\times$), \emph{parallel} ($\wedge$), and \emph{loop} ($\circlearrowleft$). 

\begin{definition}[Process Tree Syntax]
\label{def:process-tree-syntax}
Let $\mathcal{A}$ be the universe of activities with $\tau {\notin}  \mathcal{A}$. 
Let $\bigoplus {=} \{\rightarrow,\times,\wedge,\circlearrowleft\}$ be the set of process tree operators.
We define a process tree $T{=}(V,E,\lambda,r)$ consisting of a totally ordered set of nodes $V$, a set of edges $E{\subseteq}V{\times}V$, a labeling function $\lambda {:} V {\to} \mathcal{A}{\cup}\{\tau\}{\cup}\bigoplus$, and a root node $r{\in}V$. 

\begin{itemize}[noitemsep,topsep=0pt]
    \item $\big( \{n\},\{\},\lambda,n \big)$ with $\lambda(n){\in} \mathcal{A}{\cup}\{\tau\}$ is a process tree

   \item given $k{>}1$ trees $T_1{=}(V_1,E_1,\lambda_1,r_1),\dots, \allowbreak T_k{=}\allowbreak(V_k,\allowbreak E_k,\allowbreak\lambda_k,r_k)$ with $r{\notin}\allowbreak V_1{\cup}\dots{\cup}V_k$ and $\forall i,j {\in} \{1,\dots,k\}(i{\neq} j \Rightarrow V_i {\cap} V_j {=} \emptyset)$ then $\pt$ is a tree s.t.:
   \begin{itemize}[noitemsep,topsep=0pt]
       \item $V{=}V_1{\cup}\dots{\cup}V_k{\cup}\{r\}$
       \item $E{=}E_1{\cup}\dots{\cup}E_k{\cup}\big\{ (r,r_1),\dots,(r,r_k) \big\}$
       \item $\lambda(x){=}\lambda_j(x)$ for all $j{\in}\{1,\dots,k\}, x{\in}V_j$
       \item $\lambda(r){\in}\bigoplus$ and $  \lambda(r){=}{\circlearrowleft}\Rightarrow k{=} 2$
   \end{itemize}

\end{itemize}
We denote the universe of process trees by $\mathcal{T}$.

\end{definition}

\begin{figure}[tb]
    \centering
    \begin{tikzpicture}[every label/.style={text=darkgray,font=\scriptsize}]
        \tikzstyle{tree_op}=[circle,draw=black,fill=gray!30,thick,minimum size=4.5mm,inner sep=0pt]
        \tikzstyle{tree_leaf}=[rectangle,draw=black,thick,minimum size=4.5mm,inner sep=0pt]
        \tikzstyle{tree_leaf_inv}=[rectangle,draw=black,fill=black,thick,minimum size=4.5mm, text=white,inner sep=0pt]
        \tikzstyle{marking}=[dashed, draw=gray]

        \tikzstyle{level 1}=[sibling distance=38mm,level distance=4mm]
        \tikzstyle{level 2}=[sibling distance=20mm,level distance=4mm]
        \tikzstyle{level 3}=[sibling distance=20mm, level distance=5mm]
        \tikzstyle{level 4}=[sibling distance=12mm,level distance=4mm]
        
        \node [tree_op, label=$n_0$] (root){$\rightarrow $}
          child {node [tree_op, label=$n_{1.1}$] (loop) {$\circlearrowleft $}
            child { node [tree_op, label=$n_{2.1}$] (choice) {$\times$}
              child {node [tree_op, label=$n_{3.1}$] (sequence) {$\rightarrow$}
                child {node [tree_leaf, label=$n_{4.1}$] (a) {$a$}}
                child {node [tree_leaf, label=$n_{4.2}$] (b) {$b$}}
              } 
              child {node [tree_op,label=$n_{3.2}$] (parallel1) {$\wedge$}
                child {node [tree_leaf, label=$n_{4.3}$] (c) {$c$}}
                child {node [tree_leaf, label=$n_{4.4}$] (d) {$d$}}
              } 
            }
            child {node [tree_leaf_inv, label=$n_{2.2}$] (tau) {$\tau$}}
          }
          child {node [tree_op, label=$n_{1.2}$] (parallel2) {$\wedge$}
            child {node [tree_leaf, label=$n_{2.3}$] (e) {$e$}}
          	child {node [tree_leaf, label=$n_{2.4}$] (f) {$a$}}
          }
        ;
 
       \node[marking,fit=(loop) (a) (d) (tau) ,inner sep=5pt,yshift=0.1cm, label=above:{$T_1 {=} \triangle^{T_0}(n_{1.1})$}] {};
       \node[marking,fit=(parallel2) (e) (f) ,inner sep=5pt,yshift=0.1cm, label=above:{$T_2 {=} \triangle^{T_0}(n_{1.2})$}] {};
     
    \end{tikzpicture}
    
    \caption{Process tree $T_0 {=}\big(  \{n_o,\dots,n_{4.4}\},\allowbreak \big\{ (n_0,n_{1.1}),\allowbreak\dots,\allowbreak (n_{3.2},n_{4.4})\big\},\allowbreak \lambda,\allowbreak n_0\big)$ with $\lambda(n_0){=}{\rightarrow},\dots,\allowbreak\lambda(n_{4.4}){=}d$}
    
    \label{fig:process_tree_example}
\end{figure}

\noindent Note that every operator (inner node) has at least two children except for the loop operator which always has exactly two children (\autoref{def:process-tree-syntax}).
Next to the graphical representation, any process tree can be textually represented because of its totally ordered node set, e.g.,
$T_0 {\widehat{=}} {\rightarrow} \big( {\circlearrowleft} \big( {\times} \big( {\rightarrow} (a,b) ,  \allowbreak{\wedge} (c,d)    \big) ,\tau \big)   ,\allowbreak {\wedge} (e,a) \big)$. 

Given two process trees $T_1{=}(V_1,E_1,\lambda_1,r_1),\allowbreak T_2{=}(V_2,\allowbreak E_2,\lambda_2,r_2) {\in}\mathcal{T}$, we call $T_1$ a \emph{subtree} of $T_2$, written as $T_1 {\sqsubseteq} T_2$, iff $V_1 {\subseteq} V_2, E_1 {\subseteq} E_2,\allowbreak r_1{\in}V_2$, and $\forall n {\in} V_1:\lambda_1(n){=}\lambda_2(n)$.
For instance, $T_1{\sqsubseteq}T_0$ and $T_1{\not\sqsubseteq}T_2$ in \autoref{fig:process_tree_example}.

The degree indicates the number of edges connected to a node.
We distinguish between incoming $d^+$ and outgoing edges $d^-$, e.g., $d^+(n_{2.1}){=}1$ and $d^-(n_{2.1}){=}2$. 
For a tree $\pt$, we denote its \emph{leaf nodes} by $T^L{=}\{v{\in}V {\mid}\allowbreak d^-(v){=} 0\}$.
The child function $c^T {:} V {\to} V^*$ returns a sequence of child nodes according to the order of $V$, i.e., $c^T(v){=}\langle v_1,\dots,v_j\rangle$ s.t. $(v,v_1),\dots,(v,v_j){\in}E$.
For instance, $c^T_0(n_{1.1}){=}\langle n_{2.1},\allowbreak n_{2.2}\rangle$. 
For $\pt$ and a node $v{\in}V$, $\triangle^T(v)$ returns the corresponding subtree $T'$ s.t. $v$ is the root node.
Consider $T_0$, $\triangle^{T_0}(n_{1.1}){=}\allowbreak T_1$.
$p^T {:} V {\to} V {\cup} \{\perp\}$ returns the unique parent of a given node or $\perp$ for the root node.

For $\pt$ and nodes $n_1,n_2 {\in} V$, we define the \emph{lowest common ancestor (LCA)} as $LCA(n_1,n_2) {=} n{\in}V$ such that for $\triangle^T(n){=}(V_n,\allowbreak E_n,\allowbreak \lambda_n,\allowbreak r_n)$ $n_1,n_2 {\in} V_n$ and the distance (number of edges) between $n$ and $r$ is maximal.
For example, $LCA(n_{4.4},\allowbreak n_{2.2}){=}\allowbreak n_{1.1}$ and $LCA(n_{4.4},\allowbreak n_{2.3}){=}\allowbreak n_{0}$ (\autoref{fig:process_tree_example}).

Next, we define running sequences and the language of process trees.

\begin{definition}[Process Tree Running Sequences]
For the universe of activities $\mathcal{A}$ (with $\tau,open,close {\notin} \mathcal{A}$), $\pt\allowbreak{\in}\mathcal{T}$, we recursively define its running sequences $\mathcal{RS}(T){\subseteq}\big(V  {\times}  (\mathcal{A}  {\cup} \{\tau\} {\cup} \{open,close\} )\big)^*$.
\begin{itemize}[itemsep=3pt,topsep=1pt]
    \item if $ \lambda(r){\in}\mathcal{A}{\cup}\{\tau\}$ ($T$ is a leaf node):
    $\mathcal{RS}(T){=}\big\{\big\langle (r,\lambda(r))\rangle\big\}$

    \item if $\lambda(r){=}\rightarrow$ with child nodes $c^T(r){=}\langle v_1,\dots,v_k\rangle$ for $k{\geq}1$:
    \newline
    $\mathcal{RS}(T){=} \big\{\big\langle (r,open) \big\rangle\big\} {\cdot} \mathcal{RS}(\triangle^T(v_1)) {\cdot}\dots{\cdot}  \mathcal{RS}(\triangle^T(v_k)) {\cdot}\allowbreak \big\{\big\langle (r,close) \big\rangle\big\}$
 
    \item if $\lambda(r){=}\times$ with child nodes $c^T(r){=}\langle v_1,\dots,v_k\rangle$ for $k{\geq}1$:
    \newline
    $\mathcal{RS}(T){=}\big\{\big\langle (r,open) \big\rangle\big\} {\cdot} \big\{\mathcal{RS}(\triangle^T(v_1)) {\cup}\dots{\cup}  \mathcal{RS}(\triangle^T(v_k))\big\} {\cdot} \big\{\big\langle (r,close) \big\rangle\big\} $ 
         
    \item if $\lambda(r){=}\wedge$ with child nodes $c^T(r){=}\langle v_1,\dots,v_k\rangle$ for $k{\geq}1$:
    \newline
    $\mathcal{RS}(T){=} \big\{\big\langle (r,open) \big\rangle\big\} {\cdot} \big\{ \mathcal{RS}(\triangle^T(v_1)) {\diamond}\dots{\diamond} \mathcal{RS}(\triangle^T(v_k)) \big\} {\cdot} \big\{\big\langle (r,close) \big\rangle\big\}$

    \item if $\lambda(r){=}{\circlearrowleft}$  with child nodes $c^T(r){=}\langle v_1,v_2\rangle$:\newline 
    $\mathcal{RS}(T){=}\big\{ \big\langle (r,open) \big\rangle {\cdot} \sigma_1 {\cdot} \sigma_1' {\cdot} \sigma_2 {\cdot} \sigma_2' {\cdot}\allowbreak {\dots} {\cdot} \sigma_m {\cdot} \big\langle (r,close) \big\rangle \mid m {\geq} 1 \land\allowbreak \forall{ 1 {\leq} i {\leq} m} \allowbreak \big(\sigma_i{\in}\allowbreak\mathcal{RS}(\triangle^T(v_1))\big) \land\allowbreak \forall{1 {\leq} i {\leq} m{-}1} \allowbreak \big(\sigma_i'{\in}\mathcal{RS}(\triangle^T(v_2))\big) \big\}$ 
    
\end{itemize}
\end{definition}

\begin{definition}[Process Tree Language]
For given $T{\in}\mathcal{T}$, we define its language by $\mathcal{L}(T) {:=} \big\{ \big( \pi_2^*(\sigma)\big)_{\downarrow_{\mathcal{A}}} \mid \sigma{\in} \mathcal{RS}(T) \big\} {\subseteq}\mathcal{A}^*$.
\end{definition}

\noindent For example, consider the running sequences of $T_2$ (\autoref{fig:process_tree_example}), i.e., $\mathcal{RS}(T_2){=}\allowbreak \big\{\allowbreak \big\langle \allowbreak(n_{1.2},\allowbreak open),\allowbreak (n_{2.3},e),(n_{2.4},a)),(n_{1.2},close)\allowbreak \big\rangle$,\allowbreak $\big\langle (n_{1.2},open),\allowbreak (n_{2.4},a),(n_{2.3},e) ,\allowbreak (n_{1.2},\allowbreak close) \big\rangle\big\}$.
Hence, this subtree describes the language $\mathcal{L}(T_2) {=} \big\{ \langle e,a \rangle,\langle a,e \rangle \big\}$.

\subsection{Alignments}
\label{sec:alignments}

\begin{figure}[tb]
    \scriptsize
    \newcolumntype{l}{>{\columncolor{logMove}\color{white}}c}
    \newcolumntype{s}{>{\columncolor{synchronousMove}}c}
    \newcolumntype{m}{>{\columncolor{modelMove}\color{white}}c}
    \newcolumntype{i}{>{\color{black}}c}
    \centering
    \begin{tabular}{| i | i | i | s |s | i | i|i|i|i|s|m|l|i|i|i|}\hline
        $\gg$    
        & $\gg$
        & $\gg$   
        
        & $a$
        & $b$
        
        & $\gg$   
        & $\gg$ 
        & $\gg$ 
        & $\gg$ 
        & $\gg$ 
        
        & $c$
        & $\gg$ 
        & $f$

        & $\gg$ 
        & $\gg$ 
        & $\gg$
        
 \\ \hline
        
        \makecell{$(n_{1.1},$ \\ $open)$} 
        & \makecell{$(n_{2.1},$ \\ $open)$} 
        & \makecell{$(n_{3.1},$ \\ $open)$} 

        & \makecell{$(n_{4.1},$ \\ $a)$}   
        & \makecell{$(n_{4.2},$ \\ $b)$}   
        
        & \makecell{$(n_{3.1},$ \\ $close)$} 
        & \makecell{$(n_{2.1},$ \\ $close)$}
        & \makecell{$(n_{2.2},$ \\ $\tau)$}
        & \makecell{$(n_{2.1},$ \\ $open)$} 
        & \makecell{$(n_{3.2},$ \\ $open)$} 
        & \makecell{$(n_{4.3},$ \\ $c)$}   
        & \makecell{$(n_{4.4},$ \\ $d)$}   
        & $\gg$   
        
        & \makecell{$(n_{3.2},$ \\ $close)$} 
        & \makecell{$(n_{2.1},$ \\ $close)$} 
        & \makecell{$(n_{1.1},$ \\ $close)$}
\\ \hline
    \end{tabular}

    \caption{Optimal alignment $\gamma{=}\big\langle \big( \gg,(n_{1.1},open)\big),\dots,\big( \gg,(n_{1.1},close)\big) \big\rangle$ for the trace $\langle a,b,c,f \rangle$ and the process tree $T_1$ (\autoref{fig:process_tree_example})}
    \label{fig:alignments}
\end{figure}

\emph{Alignments} quantify deviations between observed process behavior (event data) and modeled behavior (process models)~\cite{DBLP:books/sp/CarmonaDSW18}.
\autoref{fig:alignments} shows an alignment for the trace $\langle a,b,c,f \rangle$ and $T_1$ (\autoref{fig:process_tree_example}).
Ignoring the skip-symbol $\gg$, the first row of an alignment always corresponds to the trace and the second row to a running sequence of the tree.
In general, we distinguish four alignment move types.
\begin{enumerate}[noitemsep,topsep=0pt]
    \item \textbf{synchronous moves} (shown \colorbox{synchronousMove}{\color{black}light-gray} in \autoref{fig:alignments}) indicate \emph{no} deviation
    \item \textbf{log moves} (shown \colorbox{black}{\color{white}black} in \autoref{fig:alignments}) indicate a \emph{deviation}, i.e., the observed activity in the trace is not executable in the model (at this point)
    \item \textbf{visible model moves} (shown \colorbox{modelMove}{\color{white}dark-gray} in \autoref{fig:alignments}) indicate a \emph{deviation}, i.e., an activity not observed in the trace must be executed w.r.t. the model 
    \item \textbf{invisible $(\tau,open,close)$ model moves} (shown \cfbox{black}{white} in \autoref{fig:alignments}) indicate \emph{no} deviation, i.e., opening or closing of a subtree or an executed $\tau$ leaf node
\end{enumerate}
\begin{definition}[Alignment]
Let $\mathcal{A}$ be the universe of activities, let $\gg,\tau {\notin} \mathcal{A}$, $\sigma {\in} \mathcal{A}^*$ be a trace and $\pt {\in} \mathcal{T}$ be a tree.
A sequence\\$\gamma {\in} \Big( \big(\mathcal{A}{\cup}\{\gg\}\big)\allowbreak \times\allowbreak \big( (V  {\times}  (\mathcal{A}  {\cup} \{\tau\} {\cup} \{open,close\} )) \cup\{\gg\} \big) \Big)^*$ is an alignment iff:
\begin{enumerate}[noitemsep,topsep=0pt]
    \item  $\sigma {=} \pi^*_1(\gamma)_{\downarrow_{\mathcal{A}}}$
    
    \item  $\pi^*_2(\gamma)_{\downarrow_{(V  {\times}  (\mathcal{A}  {\cup} \{\tau\} {\cup} \{open,close\} ))}}  {\in} \mathcal{RS}(T)$
    
    \item $(\gg,\gg) {\notin} \gamma$ and $\forall a {\in} \mathcal{A}  \forall s {\in} \mathcal{RS}(T) \big( (a,s) {\in} \gamma  \Rightarrow a {=} \lambda(\pi_2(s))   \big)$
\end{enumerate}
\end{definition}
Since multiple alignments exist for a given tree and trace, we are interested in an \emph{optimal alignment}. 
An alignment is optimal if it minimizes the deviations, i.e.,  the number of log moves and visible model moves. 
Below, we show an example.

\section{Freezing Approach}
\label{sec:freezing}
First, we introduce a formal definition of freezing-enabled incremental process discovery algorithms in \autoref{sec:overview}.
\autoref{sec:baseline_approach} introduces a baseline approach, and \autoref{sec:advanced_approach} introduces the main proposed approach.

\subsection{Problem Definition}
\label{sec:overview}

Reconsider \autoref{fig:freezing_overview} showing the overall framework of our proposal. 
A user incrementally selects subtrees from a process tree \enquote{under construction} and a trace $\sigma$ from an event log. 
Both, the tree with frozen subtree(s) and the trace, are the input for an \emph{freezing-enabled} incremental process discovery algorithm, which returns a modified tree that contains the frozen subtree(s) and accepts the selected trace. 
Next, we define a Incremental Process Discovery Algorithm (IPDA).
 

\begin{definition}[Incremental Process Discovery Algorithm]
\\
$\alpha{\colon}\mathcal{T}{\times}\mathcal{A}^*{\times}\mathcal{P}(\mathcal{A}^*){\nrightarrow}\mathcal{T}$ is an IPDA if for any tree $T{\in}\mathcal{T}$, trace $\sigma{\in}\mathcal{A}^*$, 
and previously added traces $\mathbf{P}{\in}\mathcal{P}(\mathcal{A}^*)$ with $\mathbf{P} {\subseteq} \mathcal{L}(T)$ 
it holds that  $\{\sigma\}{\cup}\mathbf{P}{\subseteq}\mathcal{L}\big(\alpha(T,\sigma,\mathbf{P})\big)$.

\noindent If $\mathbf{P} {\nsubseteq} \mathcal{L}(T)$, $\alpha$ is undefined. 

\end{definition}

\begin{figure}[tb]
    \centering
    \includegraphics[width=.7\textwidth,clip,trim=0cm 5.8cm 5.4cm 0cm]{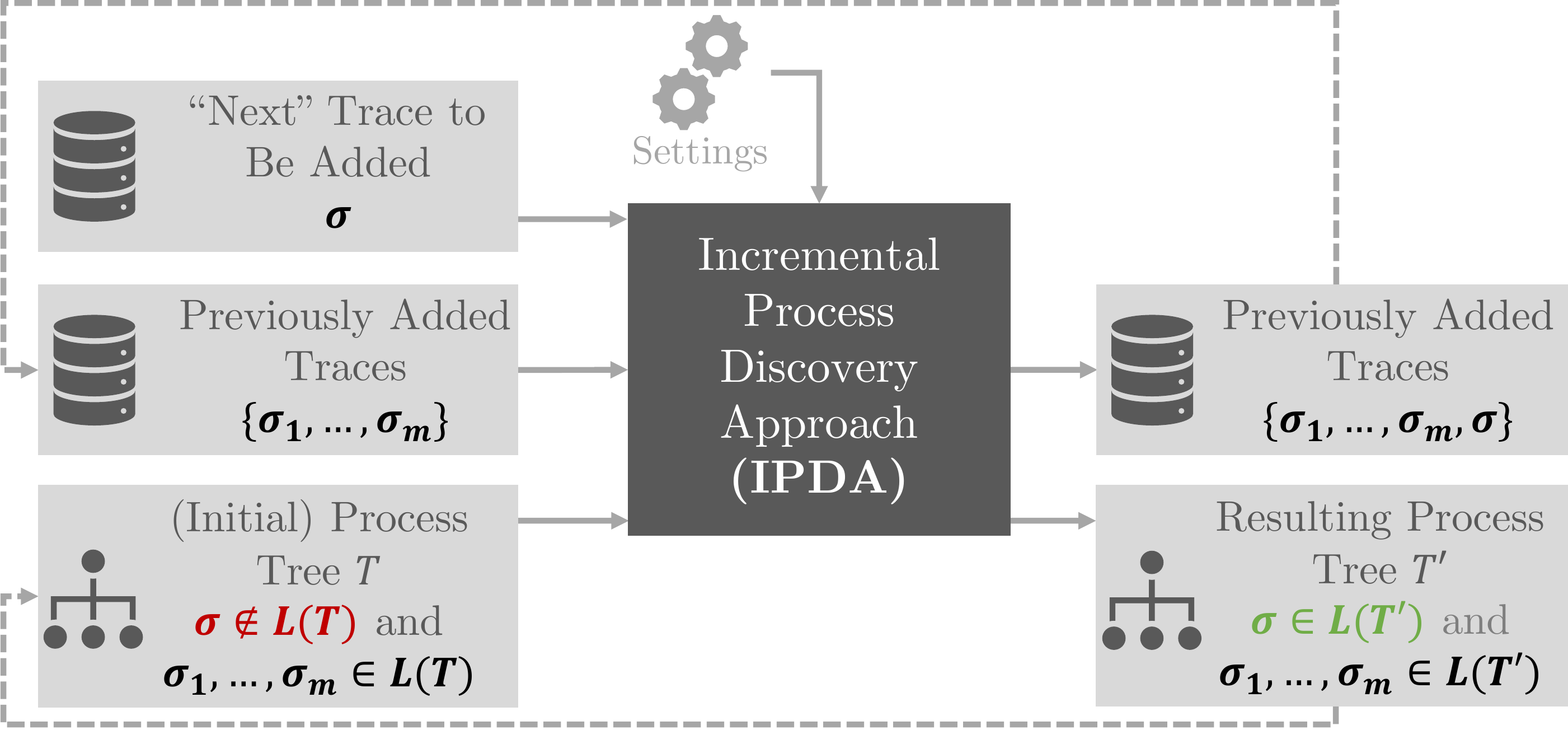}
    \caption{Schematic overview of incremental process discovery algorithms}
    \label{fig:overview_incremental}
\end{figure}

\autoref{fig:overview_incremental} shows an overview of IPDAs and their incremental application.
Starting from an (initial) tree $T$, a user incrementally selects a trace $\sigma$ not yet described by $T$.
The algorithm alters the process tree $T$ into $T'$ that accepts $\sigma$ and the previously selected/added traces. 
$T'$ is then used as input for the next incremental execution.
For a specific example of an IPDA, we refer to our previous work~\cite{10.1007/978-3-030-50316-1_25}.
Next, we formally define a freezing-enabled IPDA.

\begin{definition}[Freezing-Enabled Incremental Process Discovery Algorithm]
\label{def:freezingIPDA}
$\alpha_f{\colon}\mathcal{T}{\times}\mathcal{A}^*{\times}\mathcal{P}(\mathcal{A}^*){\times}\mathcal{P}(\mathcal{T}){\nrightarrow}\mathcal{T}$ is a freezing-enabled IPDA if for any tree $T{\in}\mathcal{T}$,
trace $\sigma{\in}\mathcal{A}^*$, 
previously added traces $\mathbf{P}{\in}\mathcal{P}(\mathcal{A}^*)$ with $\mathbf{P} {\subseteq} \mathcal{L}(T)$,
and $n{\geq}0$ frozen subtrees 
$\mathbf{T}{=}\{T_1,\allowbreak\dots,T_n\}\allowbreak{\in}\allowbreak\mathcal{P}(\mathcal{T})$ s.t. 
$\forall i,j {\in} \{1,\dots,n\} (T_i{\sqsubseteq} T \land i{\neq}j \Rightarrow T_i {\not\sqsubseteq} T_j) $ it holds that 
$\{\sigma\}{\cup}\mathbf{P}{\subseteq}\mathcal{L}\big(\alpha_f(\allowbreak T,\sigma,\mathbf{P},\mathbf{T})\big)$ 
and $\forall{T'{\in}\mathbf{T}} \big(T'{\sqsubseteq}\alpha_f(T,\sigma,\mathbf{P},\mathbf{T})\big)$.

\noindent If $\mathbf{P} {\nsubseteq} \mathcal{L}(T)$ or $\exists i,j {\in} \{1,\dots,n\} (T_i{\not\sqsubseteq} T \lor i{\neq}j \Rightarrow T_i {\sqsubseteq} T_j) $, $\alpha_f$ is undefined. 

\end{definition}

In \autoref{sec:baseline_approach} and \autoref{sec:advanced_approach}, we present two freezing-enabled IPDAs, i.e., instantiations of $\alpha_f$ according to \autoref{def:freezingIPDA}.

\subsection{Baseline Approach}
\label{sec:baseline_approach}

This section presents a baseline approach, i.e., a freezing-enabled IPDA.
Consider \autoref{alg:baseline_freezing}.
The central idea is to apply an IPDA ignoring the frozen subtrees (\autoref{alg:baseline:line:incr}).
Next, we check if the returned process tree contains the frozen subtree(s).
If this is the case, we return the process tree $T'$ (\autoref{alg:baseline:line:return1}).
Otherwise, we put the altered tree $T'$ in parallel with all frozen subtrees which are not contained in $T'$ and make these frozen subtrees optional (\autoref{alg:baseline:line:return2}). 

\begin{algorithm}[tb]
    
    \scriptsize
	
	\caption{Baseline Approach Freezing Subtrees}
	\label{alg:baseline_freezing}
	\SetKwInOut{Input}{input}
	\Input{$\pt{\in}\mathcal{T}, 
	T_1,\dots,T_n {\sqsubseteq} T (n {\geq} 0), 
	\sigma{\in}\mathcal{A}^*, 
	 \sigma_1,\dots,\sigma_m{\in}\mathcal{A}^*$}
	\Begin{
	    
	    \nl $T' \gets \alpha(T,\sigma, \{T_1,\dots,T_n\})$\tcp*[r]{apply a non-freezing-enabled IPDA} \label{alg:baseline:line:incr}
	    
	    \nl $\mathbf{T_1} \gets \big\{T_i \mid T_i {\in} \{T_1,\dots,T_n\} \land T_i {\sqsubseteq} T' \big\}$\;
	    
	    \nl $\mathbf{T_2} \gets \big\{T_i \mid T_i {\in} \{T_1,\dots,T_n\} \land T_i {\not\sqsubseteq} T' \big\}$\;

	    \nl \If(\tcp*[h]{all frozen subtrees $T_1,\dots,T_n$ are contained in $T'$}){$\mathbf{T_2}{=}\emptyset$}{
	        \nl \Return $T'$\; \label{alg:baseline:line:return1}
	    }
	    \nl \Else(){
	      
	        \nl \Return ${\wedge}\big(T',{\times}(T_{i_1},\tau),\dots,{\times}(T_{i_j},\tau)\big)$ for $T_{i_1},\dots,T_{i_j}{\in}\mathbf{T_2}$\;\label{alg:baseline:line:return2}

	    }
	}			
\end{algorithm}

For example, assume the process tree $T_0$ with frozen subtree $T_2$ (\autoref{fig:process_tree_example}).
The next trace to be added is $\sigma {=} \langle c,d,a,e,a,a,e \rangle$ and the set of previously added traces is $\{\sigma_1{=}\langle d,c,a,b,a,e \rangle,\sigma_2{=}\langle a,b,e,a \rangle\}$.
Applying \autoref{alg:baseline_freezing} could return the tree $T' {\widehat{=}} {\rightarrow} \big({\wedge} \big( {\times} \big( {\rightarrow} (a,b) ,  \allowbreak{\wedge} (c,d)    \big) ,\tau \big),\allowbreak {\wedge}(e, {\circlearrowleft}(a,\tau)) \big) $ (at \autoref{alg:baseline:line:incr}) depending on the specific choice of $\alpha$.
$T'$ allows for multiple executions of $a$ in the end compared to $T$; thus, $\{\sigma,\sigma_1,\sigma_2\} {\subseteq} \mathcal{L}(T')$.
However, $T'$ does not contain the frozen subtree $T_2$ anymore.
Hence, we put $T'$ in parallel with the frozen subtree $T_2$.
Finally, we return ${\wedge}\big( {\rightarrow} \big({\wedge} \big( {\times} \big( {\rightarrow} (a,b) ,  \allowbreak{\wedge} (c,d)    \big) ,\tau \big),\allowbreak {\wedge}(e, {\circlearrowleft}(a,\tau)) \big), {\times}(\tau,{\wedge (e,a) )} \big)$ (\autoref{alg:baseline:line:return2}).

\subsection{Advanced Approach}
\label{sec:advanced_approach}

\begin{figure}[b]
    \centering
    \includegraphics[width=\textwidth,trim=0cm 2.7cm 15.5cm 0cm,clip]{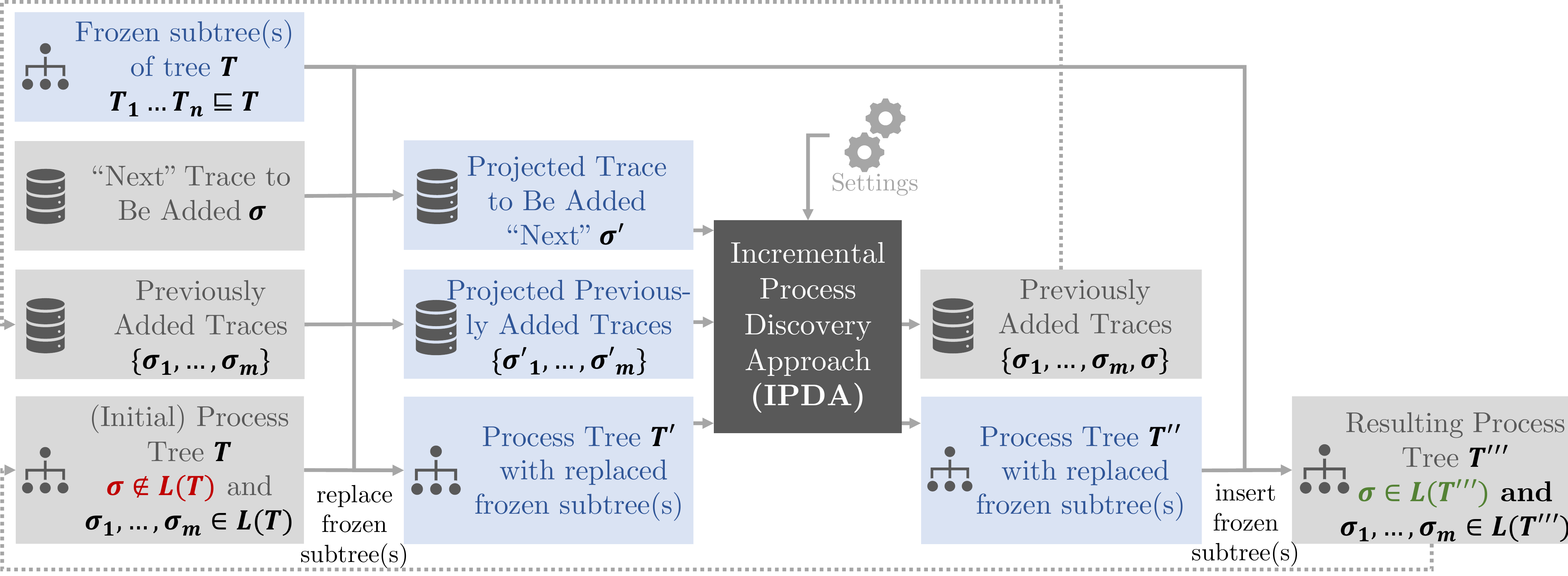}
    \caption{Overview of the proposed freezing-enabled IPDA
    }
    \label{fig:freezing_advanced}
\end{figure}

\begin{figure}[t]
    \centering
    \begin{subfigure}[b]{0.49\textwidth}
        \centering
        \begin{tikzpicture}[every label/.style={text=darkgray,font=\scriptsize}]
            \tikzstyle{tree_op}=[circle,draw=black,fill=gray!30,thick,minimum size=4.5mm,inner sep=0pt]
            \tikzstyle{tree_leaf}=[rectangle,draw=black,thick,minimum size=4.5mm,inner sep=0pt]
            \tikzstyle{tree_leaf_inv}=[rectangle,draw=black,fill=black,thick,minimum size=4.5mm, text=white,inner sep=0pt]
            \tikzstyle{marking}=[dashed, draw=gray]
    
            \tikzstyle{level 1}=[sibling distance=25mm,level distance=3mm]
            \tikzstyle{level 2}=[sibling distance=9mm,level distance=5.5mm]
            \tikzstyle{level 3}=[sibling distance=12mm, level distance=5.5mm]
            \tikzstyle{level 4}=[sibling distance=6mm,level distance=5.5mm]
            
            \node [tree_op] (root){$\rightarrow $}
              child {node [tree_op, ] (loop) {$\circlearrowleft $}
                child { node [tree_op, ] (choice) {$\times$}
                  child {node [tree_op, ] (sequence) {$\rightarrow$}
                    child {node [tree_leaf, ] (a) {$a$}}
                    child {node [tree_leaf, ] (b) {$b$}}
                  } 
                  child {node [tree_op,] (parallel1) {$\wedge$}
                    child {node [tree_leaf, ] (c) {$c$}}
                    child {node [tree_leaf, ] (d) {$d$}}
                  } 
                }
                child {node [tree_leaf_inv] (tau) {$\tau$}}
              }
              child {node [tree_op, ] (parallel2) {$\wedge$}
                child {node [tree_leaf,] (e) {$e$}}
              	child {node [tree_leaf] (f) {$a$}}
              }
            ;
     
           \node[marking,fit=(parallel2) (e) (f) ,inner sep=3pt,yshift=0cm, label=below:{frozen subtree $T_2$}] {};
         
        \end{tikzpicture}        
        \caption{Initial tree $T$ (same as shown in \autoref{fig:process_tree_example}) with frozen subtree $T_2$}
        \label{fig:running-example-a}
    \end{subfigure}
    \hfill
    \begin{subfigure}[b]{0.49\textwidth}
        \centering
        \begin{tikzpicture}[every label/.style={text=darkgray,font=\scriptsize}]
            \tikzstyle{tree_op}=[circle,draw=black,fill=gray!30,thick,minimum size=4.5mm,inner sep=0pt]
            \tikzstyle{tree_leaf}=[rectangle,draw=black,thick,minimum size=4.5mm,inner sep=0pt]
            \tikzstyle{tree_leaf_inv}=[rectangle,draw=black,fill=black,thick,minimum size=4.5mm, text=white,inner sep=0pt]
            \tikzstyle{marking}=[dashed, draw=gray]
    
            \tikzstyle{level 1}=[sibling distance=25mm,level distance=3mm]
            \tikzstyle{level 2}=[sibling distance=9mm,level distance=5.5mm]
            \tikzstyle{level 3}=[sibling distance=12mm, level distance=5.5mm]
            \tikzstyle{level 4}=[sibling distance=6mm,level distance=5.5mm]
            
            \node [tree_op] (root){$\rightarrow $}
              child[sibling distance=15mm] {node [tree_op, ] (loop) {$\circlearrowleft $}
                child[sibling distance=6mm] { node [tree_op, ] (choice) {$\times$}
                  child {node [tree_op, ] (sequence) {$\rightarrow$}
                    child {node [tree_leaf, ] (a) {$a$}}
                    child {node [tree_leaf, ] (b) {$b$}}
                  } 
                  child {node [tree_op,] (parallel1) {$\wedge$}
                    child {node [tree_leaf, ] (c) {$c$}}
                    child {node [tree_leaf, ] (d) {$d$}}
                  } 
                }
                child[sibling distance=6mm] {node [tree_leaf_inv] (tau) {$\tau$}}
              }
              child[sibling distance=30mm] {node [tree_op, ] (parallel2) {$\rightarrow$}
                child[sibling distance=11mm] {node [tree_leaf,] (e) {$open^{T_2}$}}
              	child[sibling distance=11mm] {node [tree_leaf] (f) {$close^{T_2}$}}
              }
            ;
     
           \node[marking,fit=(parallel2) (e) (f) ,inner sep=3pt,yshift=0cm, label=below:{replaced frozen subtree}] {};
         
        \end{tikzpicture}  
        \caption{Tree $T'$ with replaced frozen subtree}
        \label{fig:running-example-b}
    \end{subfigure}
    \hfill
    \begin{subfigure}[b]{0.49\textwidth}
        \centering
        \begin{tikzpicture}[every label/.style={text=darkgray,font=\scriptsize}]
            \tikzstyle{tree_op}=[circle,draw=black,fill=gray!30,thick,minimum size=4.5mm,inner sep=0pt]
            \tikzstyle{tree_leaf}=[rectangle,draw=black,thick,minimum size=4.5mm,inner sep=0pt]
            \tikzstyle{tree_leaf_inv}=[rectangle,draw=black,fill=black,thick,minimum size=4.5mm, text=white,inner sep=0pt]
            \tikzstyle{marking}=[dashed, draw=gray]
            
            \tikzstyle{level 1}=[sibling distance=25mm,level distance=3mm]
            \tikzstyle{level 2}=[sibling distance=9mm,level distance=5.5mm]
            \tikzstyle{level 3}=[sibling distance=12mm, level distance=5.5mm]
            \tikzstyle{level 4}=[sibling distance=6mm,level distance=5.5mm]
            
            \node [tree_op] (root){$\rightarrow $}
              child {node [tree_op, ] (loop) {$\circlearrowleft $}
                child[sibling distance=6mm] { node [tree_op, ] (choice) {$\times$}
                  child {node [tree_op, ] (sequence) {$\rightarrow$}
                    child {node [tree_leaf, ] (a) {$a$}}
                    child {node [tree_leaf, ] (b) {$b$}}
                  } 
                  child {node [tree_op,] (parallel1) {$\wedge$}
                    child {node [tree_leaf, ] (c) {$c$}}
                    child {node [tree_leaf, ] (d) {$d$}}
                  } 
                }
                child[sibling distance=6mm] {node [tree_leaf_inv] (tau) {$\tau$}}
              }
              child {node [tree_op, ] (parallel2) {$\rightarrow$}
                child[sibling distance=8mm] {node [tree_leaf,] (e) {$open^{T_2}$}}
                child[sibling distance=8mm] { node [tree_op, ] (choice2) {$\circlearrowleft$}
                    child[sibling distance=6mm] {node [tree_leaf_inv] (tau2) {$\tau$}}
                    child[sibling distance=6mm] {node [tree_leaf, ] (a2) {$a$}}
                }
              	child[sibling distance=8mm] {node [tree_leaf] (f) {$close^{T_2}$}}
              }
            ;
     
         
        \end{tikzpicture}
        \caption{Tree $T''$ after applying an IPDA}
        \label{fig:running-example-c}
    \end{subfigure}
    \begin{subfigure}[b]{0.49\textwidth}
        \centering
        \begin{tikzpicture}[every label/.style={text=darkgray,font=\scriptsize}]
            \tikzstyle{tree_op}=[circle,draw=black,fill=gray!30,thick,minimum size=4.5mm,inner sep=0pt]
            \tikzstyle{tree_leaf}=[rectangle,draw=black,thick,minimum size=4.5mm,inner sep=0pt]
            \tikzstyle{tree_leaf_inv}=[rectangle,draw=black,fill=black,thick,minimum size=4.5mm, text=white,inner sep=0pt]
            \tikzstyle{marking}=[dashed, draw=gray]
            \tikzstyle{level 1}=[sibling distance=25mm,level distance=3mm]
            \tikzstyle{level 2}=[sibling distance=9mm,level distance=5.5mm]
            \tikzstyle{level 3}=[sibling distance=12mm, level distance=5.5mm]
            \tikzstyle{level 4}=[sibling distance=6mm,level distance=5.5mm]
            
            \node [tree_op] (root){$\rightarrow $}
              child {node [tree_op, ] (loop) {$\circlearrowleft $}
                child[sibling distance=6mm] { node [tree_op, ] (choice) {$\times$}
                  child {node [tree_op, ] (sequence) {$\rightarrow$}
                    child {node [tree_leaf, ] (a) {$a$}}
                    child {node [tree_leaf, ] (b) {$b$}}
                  } 
                  child {node [tree_op,] (parallel1) {$\wedge$}
                    child {node [tree_leaf, ] (c) {$c$}}
                    child {node [tree_leaf, ] (d) {$d$}}
                  } 
                }
                child[sibling distance=6mm] {node [tree_leaf_inv] (tau) {$\tau$}}
              }
              child {node [tree_op, ] (parallel2) {$\wedge$}
                child[sibling distance=13mm] {node [tree_op, ] (parallel3) {$\wedge$}
                    child[sibling distance=6mm] {node [tree_leaf,] (e) {$e$}}
              	    child[sibling distance=6mm] {node [tree_leaf] (f) {$a$}}
                }
                child[sibling distance=13mm] { node [tree_op, ] (choice2) {$\circlearrowleft$}
                    child[sibling distance=6mm] {node [tree_leaf_inv] (tau2) {$\tau$}}
                    child[sibling distance=6mm] {node [tree_leaf, ] (a2) {$a$}}
                }
              }
            ;
     
           \node[marking,fit=(parallel3) (e) (f) ,inner sep=3pt,yshift=0cm, label={[align=center, below,yshift=-12mm]inserted \\ frozen \\ subtree $T_2$}] {};
         
        \end{tikzpicture}
        \caption{Tree $T'''$ containing frozen subtree $T_2$}
        \label{fig:running-example-d}
    \end{subfigure}
    
    \caption{Running example of the advanced freezing approach}
    \label{fig:running-example}
\end{figure}
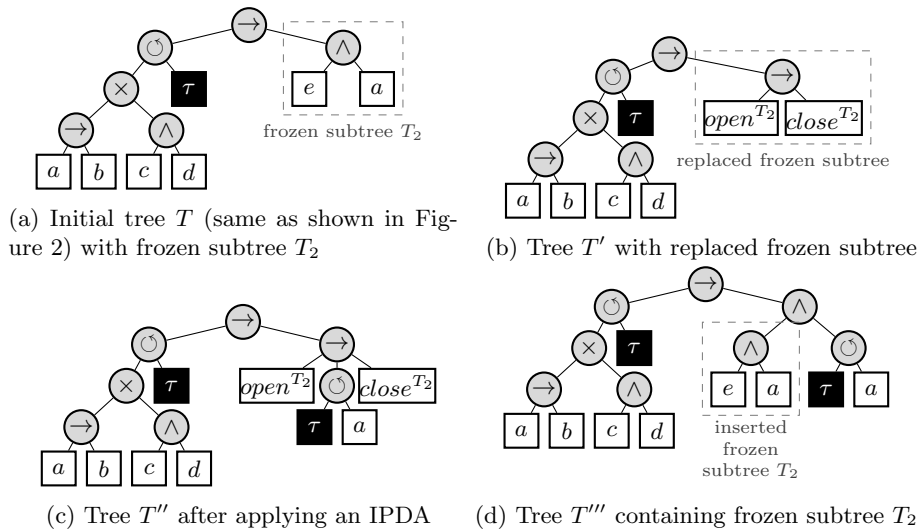

This section presents the main proposed approach that is based on an arbitrary non-freezing-enabled IPDA.
The proposed advanced freezing approach is an extension that essentially modifies the input and output artifacts of an IPDA (compare to \autoref{fig:overview_incremental}).
\autoref{fig:freezing_advanced} provides an overview of this extension.

The central idea is to replace each frozen subtree in the process tree $T$ by a new label, resulting in a modified tree $T'$ (\autoref{fig:freezing_advanced}).
Next, the previously added traces are projected, i.e., we detect full executions of the frozen subtree(s) within the traces and replace the respective activities with the corresponding new label, which is also used to replace the frozen subtree in $T$.
After applying an IPDA (visualized by the dark box in the middle of \autoref{fig:freezing_advanced}), we insert the frozen subtrees that got replaced back into the modified tree.
The remainder of this section is structured along with the input/output modifications (\autoref{fig:freezing_advanced}).

\subsubsection{Replacing Frozen Subtrees}
\label{sec:replacing_frozen_subtree}
As shown in \autoref{fig:freezing_advanced}, we use an (initial) tree $T$ and frozen subtrees $T_1,\dots,T_n {\sqsubseteq} T$ to be replaced and return a modified tree $T'$.
For example, consider \autoref{fig:running-example-a}.
We assume the tree $T$ (same as in \autoref{fig:process_tree_example}) with the frozen subtree $T_2{\widehat{=}}{\wedge}(e,a)$.
To replace the frozen subtree, we choose two unique, arbitrary labels which are not contained in the current event log nor in the tree, e.g., $open^{T_2}$ and $close^{T_2}$. 
In the remainder, we denote the universe of replacement labels by $\mathcal{R}$, e.g., $open_{T_2}{\in}\mathcal{R}$.
Now, we replace the frozen subtree ${\wedge}(e,a)$ by ${\rightarrow}(open^{T_2},close^{T_2})$ and get the resulting tree $T'$, see \autoref{fig:running-example-b}.
Semantically, $open^{T_2}$ represents the opening of the frozen subtree and $close^{T_2}$ the closing.
In general, we iteratively replace each frozen subtree.

\subsubsection{Projecting Previously Added Traces}
The set of previously added traces $\{\sigma_1,\dots,\allowbreak\sigma_m\}$ (\autoref{fig:freezing_advanced}), which fits the tree $T$, does not fit $T'$ because of the replaced frozen subtree(s).
Thus, we have to modify the traces accordingly.

We replay each previously added trace $\{\sigma_1,\dots,\sigma_m\}$ (\autoref{fig:freezing_advanced}) on $T$ and mark when a frozen subtree is \emph{opened} and \emph{closed}.
Next, we insert at all opening and closing positions the corresponding replacement label and remove all activities in between that are replayed in the corresponding frozen subtree.
Activities which are not replayed in a frozen subtree remain unchanged.

For example, reconsider $T$ (\autoref{fig:running-example-a}) and its frozen subtree $T_2$ that was replaced by ${\rightarrow}(open^{T_2},close^{T_2})$ (\autoref{fig:running-example-b}).
Assume the traces $\{\sigma_1{=}\langle d,\allowbreak c,\allowbreak a,\allowbreak b,\allowbreak a,e \rangle,\allowbreak \sigma_2{=}\langle a,b,e,a \rangle\}$.
Below, we depict the running sequence of $\sigma_1$ on $T$ and the projected trace $\sigma_1'$. 
Note that $n_{2.4}$ and $n_{2.3}$ are nodes of frozen $T_2$ (\autoref{fig:process_tree_example}).
\begin{center}
\scriptsize  

\begin{tabular}{|cccccccccc|}
\hline
\multicolumn{10}{|l|}{\color{gray}\textbf{extract of the running sequence for $\sigma_1$ on $T{=}T_0$ (see \autoref{fig:process_tree_example}):}}\\

$\langle\dots (n_{4.4},d)$,   & $(n_{4.3},c)$ & $\dots$ & $(n_{4.1},a)$,   & $(n_{4.2},b)\dots$   & \color{blue}$(n_{1.2},open)$,    & \color{red}$(n_{2.4},a)$,   & \color{red}$(n_{2.3},e)$,   &   \color{blue}$(n_{1.2},close)$   &$\dots\rangle$\\ \hline\hline


\multicolumn{10}{|l|}{\color{gray}\textbf{projected trace $\sigma_1'$ based on above running sequence:}}\\ 

$\langle d$,   & $c$, &  & $a$,   & $b$,   & \color{blue}$open^{T_2}$,    &    &    &   \color{blue}$close^{T_2}\rangle$   &  \\ \hline

\end{tabular}
\end{center}
We transform $\sigma_1{=}\langle d,c,a,b,a,e \rangle$ into $\sigma_1'{=}\langle d,\allowbreak c,\allowbreak a,\allowbreak b,\allowbreak open^{T_2}, \allowbreak close^{T_2}\rangle$ (and $\sigma_2$ into $\sigma_2' {=} \langle a,b,open^{T_2},\allowbreak close^{T_2} \rangle$).
Note that $\sigma_1',\sigma_2' {\in} \mathcal{L}(T')$ since $\sigma_1,\sigma_2 {\in} \mathcal{L}(T)$.

\subsubsection{Projecting Trace to Be Added Next}

\begin{figure}[b]
    \centering
    \begin{subfigure}[]{1\textwidth}
        \centering
        \begin{tikzpicture}[every label/.style={text=darkgray,font=\scriptsize}]
            \tikzstyle{tree_op}=[circle,draw=black,fill=gray!30,thick,minimum size=4.5mm,inner sep=0pt]
            \tikzstyle{tree_leaf}=[rectangle,draw=black,thick,minimum size=4.5mm,inner sep=0pt]
            \tikzstyle{tree_leaf_inv}=[rectangle,draw=black,fill=black,thick,minimum size=4.5mm, text=white,inner sep=0pt]
            \tikzstyle{marking}=[dashed, draw=gray]
    
            \tikzstyle{level 1}=[sibling distance=38mm,level distance=4mm]
            \tikzstyle{level 2}=[sibling distance=20mm,level distance=4mm]
            \tikzstyle{level 3}=[sibling distance=20mm, level distance=5mm]
            \tikzstyle{level 4}=[sibling distance=12mm,level distance=4mm]
            
            \node [tree_op, label=$n_0$] (root){$\circlearrowleft $}
              child {node [tree_leaf_inv, label=$n_{1.1}$] (loop) {$\tau $}
              }
              child {node [tree_op, label=$n_{1.2}$] (parallel2) {$\wedge$}
                child {node [tree_leaf, label=$n_{2.1}$] (e) {$e$}}
              	child {node [tree_leaf, label=$n_{2.2}$] (f) {$a$}}
              }
            ;
            \node[marking,fit=(parallel2) (e) (f) ,inner sep=7pt,yshift=0.1cm, label={[align=center]corresponds to\\ frozen subtree $T_2$}] {};
        \end{tikzpicture}
        \caption{Abstraction tree $A$ used to detect full executions of frozen tree $T_2$ (\autoref{fig:process_tree_example})}
        \label{fig:abstraction_tree_example}
    \end{subfigure}
    \begin{subfigure}[]{\textwidth}
        \centering
        {
    \scriptsize
    \newcolumntype{l}{>{\columncolor{logMove}\color{white}}c}
    \newcolumntype{s}{>{\columncolor{synchronousMove}}c}
    \newcolumntype{m}{>{\columncolor{modelMove}\color{white}}c}
    \newcolumntype{i}{>{\color{black}}c}
    \centering
        \begin{tabular}{|c|| l | l | i | i |s | s | i|i|i|s| l | s | i | i|}\hline
            \rowcolor{white} \color{gray}{\emph{ move index}}
            & \color{gray}{{1}} 
            & \color{gray}{{2}} 
            & \color{gray}{{3}}
            & \color{gray}{{4}} 
            & \color{gray}{{5}} 
            & \color{gray}{{6}} 
            & \color{gray}{{7}} 
            & \color{gray}{{8}} 
            & \color{gray}{{9}} 
            & \color{gray}{{10}} 
            & \color{gray}{{11}} 
            & \color{gray}{{12}} 
            & \color{gray}{{13}} 
            & \color{gray}{{14}} \\ \hline\hline
            
            \color{gray}{{\emph{trace}}} 
            & $c$    
            & $d$
            & $\gg$   
            
            & $\gg$
            & $a$
            
            & $e$   
            & $\gg$ 
            & $\gg$ 
            & $\gg$ 
            & $a$ 
            & $a$
            & $e$
            & $\gg$
            & $\gg$
            
            \\ \hline
            
            \color{gray}{{\emph{model}}} 
            & \makecell{$\gg$} 
            & \makecell{$\gg$} 
            & \makecell{$(n_{0},$ \\ $open)$} 
    
            & \makecell{$(n_{1.2},$ \\ $open)$}   
            & \makecell{$(n_{2.2},$ \\ $a)$}   
            
            & \makecell{$(n_{2.1},$ \\ $e)$} 
            & \makecell{$(n_{1.2},$ \\ $close)$}
            & \makecell{$(n_{1.1},$ \\ $\tau)$}
            & \makecell{$(n_{1.2},$ \\ $open)$} 
            & \makecell{$(n_{2.2},$ \\ $a)$} 
            & $\gg$
            & \makecell{$(n_{2.1},$ \\ $e)$}
            & \makecell{$(n_{1.2},$ \\ $close)$}
            & \makecell{$(n_{0},$ \\ $close)$} 
            \\ \hline
        \end{tabular}
        }
        \caption{Optimal alignment of $\sigma{=}\langle c,d,a,e,a,a,e \rangle$ and abstraction tree $A$}
        \label{fig:alignment_projecting_trace_to_be_added_next}
    \end{subfigure}
    \caption{Detecting full executions of the frozen subtree $T_2$ (\autoref{fig:process_tree_example}) in the trace to be added next $\sigma{=}\langle c,d,a,e,a,a,e \rangle$}
    \label{fig:three graphs}
\end{figure}

The central idea is to detect complete executions of the frozen subtree(s) within the trace.
These complete executions are then replaced by the corresponding replacement label of the frozen subtree, i.e., the activities belonging to the frozen subtree are removed and instead the $open$ and $close$ replacement label is inserted. 
Activities that are not part of a full execution of a frozen subtree remain unchanged.

Reconsider the running example (\autoref{fig:running-example}) and assume that the trace to be added next $\sigma{=}\langle c,d,a,e,a,a,e \rangle$.
To detect full executions of the frozen subtree $T_2\widehat{{=}}{\wedge}(e,a)$ independent from the entire tree $T$, we align the trace $\sigma$ with the abstraction tree $A{\widehat{=}}{\circlearrowleft}(\tau, {\wedge}(e,a))$, visualized in \autoref{fig:abstraction_tree_example}.
\autoref{fig:alignment_projecting_trace_to_be_added_next} shows an optimal alignment for $A$ and $\sigma$.
We see in the alignment that the frozen subtree $T_2$ is twice fully executed, i.e, between \nth{4} to \nth{7} and \nth{9} to \nth{13} move.
Given the alignment, we project $\sigma$ onto $\sigma' {=} \langle c,d,open^{T_2},close^{T_2},open^{T_2},a, close^{T_2} \rangle$. 

\subsubsection{Reinserting Frozen Subtrees}
This section describes how the frozen subtree(s) are reinserted into the tree $T''$, returned by the IPDA (\autoref{fig:freezing_advanced}).
Note that $T''$ can contain the same replacement label for open and close multiple times because the IPDA may add multiple leaf nodes having the same label.
Thus, we have to find appropriate position(s) in $T''$ to insert the frozen subtree(s) back.

\begin{figure}[tb]
    \scriptsize
     \centering
     \begin{subfigure}[b]{0.12\textwidth}
        \centering
        \begin{tikzpicture}
            \tikzset{
            tree_op/.style={rectangle,draw=black,fill=gray!30,thick,minimum size=4.5mm,inner sep=0pt},
            unknown_op/.style={circle,inner sep=-1.5pt},
            tree_subtree/.style={rectangle,draw = gray!15,fill=gray!15,thick,minimum size=4.5mm,inner sep=0pt},
            triangle/.style={inner sep=0pt,draw=gray!30, regular polygon, regular polygon sides=3, align=center,equal size=T,fill=gray!30, minimum height=4.5em},
            marking/.style={draw=black, densely dashed,thin},
            selected/.style={fill=darkgray, text=white, draw=black},
            level 1/.style={sibling distance=16mm,level distance=12mm},
            level 2/.style={sibling distance=16mm,level distance=8mm},
            level 3/.style={sibling distance=20mm},
            level 4/.style={sibling distance=11mm,level distance=11mm},
            equal size/.style={execute at begin
             node={\setbox\nodebox=\hbox\bgroup},
             execute at end
             node={\egroup\eqmakebox[#1][c]{\copy\nodebox}}}
            }
            
            \node 
            [unknown_op] (parent){$\bullet$}
            [child anchor=north]
                 child {node [triangle] (t1) {}}
                ;
            \node [yshift=2.25em] (lca) at (t1) {$\bullet$};
            \node[left of = lca, color=gray,xshift=.6cm](){$r_{c}$};
            \node[above of = parent, color=gray,yshift=-.8cm](label1){parent of $r_c$};

            \node[]() at (t1) {$T_{c}$};
        \end{tikzpicture}
        \caption{Initial situation}
        \label{fig:insert-frozen-subtree-back:a}
     \end{subfigure}
     \hfill
     \begin{subfigure}[b]{0.2\textwidth}
        \centering
        \begin{tikzpicture}
            \tikzstyle{tree_leaf_inv}=[circle,draw=black,fill=black,thick,minimum size=6mm, text=white,inner sep=0pt,font=\large]

            \tikzset{
            tree_op/.style={rectangle,draw=black,fill=gray!30,thick,minimum size=4mm,inner sep=0pt},
            unknown_op/.style={circle,inner sep=-1.5pt},
            tree_subtree/.style={rectangle,draw = gray!15,fill=gray!15,thick,minimum size=4.5mm,inner sep=0pt},
            triangle/.style={inner sep=0pt,draw=gray!30, regular polygon, regular polygon sides=3, align=center,equal size=T,fill=gray!30, minimum height=4.5em},
            marking/.style={draw=black, densely dashed,thin},
            selected/.style={fill=darkgray, text=white, draw=black},
            level 1/.style={sibling distance=1mm,level distance=5mm},
            level 2/.style={sibling distance=12mm,level distance=11mm},
            level 3/.style={sibling distance=15mm},
            level 4/.style={sibling distance=11mm,level distance=11mm},
            equal size/.style={execute at begin
             node={\setbox\nodebox=\hbox\bgroup},
             execute at end
             node={\egroup\eqmakebox[#1][c]{\copy\nodebox}}}
            }
            
            \node 
            [unknown_op] (parent){$\bullet$}
            [child anchor=north]
                child {node [tree_op] (parallel) {$\wedge$}
                    child {node [triangle] (t1) {}}
                    child {node [triangle,selected] (frozen) {$T_i$}}
                }
                ;
            \node [yshift=2.25em] (lca) at (t1) {$\bullet$};
            \node[left of = lca, color=gray,xshift=.6cm](){$r_c$};
            \node[above of = parent, color=gray,yshift=-.8cm](){old parent of $r_c$};
            \node[]() at (t1) {$T_{c}'$};
        \end{tikzpicture}

         \caption{Case $\{1\}$}
         \label{fig:insert-frozen-subtree-back:b}
     \end{subfigure}
     \hfill
     \begin{subfigure}[b]{0.2\textwidth}
        \centering
        \begin{tikzpicture}
            \tikzstyle{tree_leaf_inv}=[circle,draw=black,fill=black,thick,minimum size=4mm, text=white,inner sep=0pt,font=\large]

            \tikzset{
            tree_op/.style={rectangle,draw=black,fill=gray!30,thick,minimum size=4mm,inner sep=0pt},
            unknown_op/.style={circle,inner sep=-1.5pt},
            tree_subtree/.style={rectangle,draw = gray!15,fill=gray!15,thick,minimum size=4.5mm,inner sep=0pt},
            triangle/.style={inner sep=0pt,draw=gray!30, regular polygon, regular polygon sides=3, align=center,equal size=T,fill=gray!30, minimum height=4.5em},
            marking/.style={draw=black, densely dashed,thin},
            selected/.style={fill=darkgray, text=white, draw=black},
            level 1/.style={sibling distance=16mm,level distance=5mm},
            level 2/.style={sibling distance=9mm,level distance=10mm},
            level 3/.style={sibling distance=7mm},
            level 4/.style={sibling distance=11mm,level distance=11mm},
            equal size/.style={execute at begin
             node={\setbox\nodebox=\hbox\bgroup},
             execute at end
             node={\egroup\eqmakebox[#1][c]{\copy\nodebox}}}
            }
            
            \node 
            [unknown_op] (parent){$\bullet$}
            [child anchor=north]
                child {node [tree_op] (parallel) {$\wedge$}
                    child {node [triangle] (t1) {}}
                    child {node [tree_op] (choice) {$\times$}
                        child {node [tree_leaf_inv] (tau) {$\tau$}}
                        child {node [triangle, selected] (frozen) {$T_i$}}
                    }
                }
                ;
            \node [yshift=2.25em] (lca) at (t1) {$\bullet$};
            \node[left of = lca, color=gray,xshift=.6cm](){$r_c$};
            \node[above of = parent, color=gray,yshift=-.8cm](){old parent of $r_c$};
            \node[]() at (t1) {$T_{c}'$};
        \end{tikzpicture}

         \caption{Case $\{0,1\}$}
         \label{fig:insert-frozen-subtree-back:c}
     \end{subfigure}
     \hfill
     \begin{subfigure}[b]{0.2\textwidth}
         \centering
        \begin{tikzpicture}
            \tikzstyle{tree_leaf_inv}=[circle,draw=black,fill=black,thick,minimum size=4mm, text=white,inner sep=0pt,font=\large]

            \tikzset{
            tree_op/.style={rectangle,draw=black,fill=gray!30,thick,minimum size=4mm,inner sep=0pt},
            unknown_op/.style={circle,inner sep=-1.5pt},
            tree_subtree/.style={rectangle,draw = gray!15,fill=gray!15,thick,minimum size=4.5mm,inner sep=0pt},
            triangle/.style={inner sep=0pt,draw=gray!30, regular polygon, regular polygon sides=3, align=center,equal size=T,fill=gray!30, minimum height=4.5em},
            marking/.style={draw=black, densely dashed,thin},
            selected/.style={fill=darkgray, text=white, draw=black},
            level 1/.style={sibling distance=16mm,level distance=5mm},
            level 2/.style={sibling distance=9mm,level distance=10mm},
            level 3/.style={sibling distance=7mm},
            level 4/.style={sibling distance=11mm,level distance=11mm},
            equal size/.style={execute at begin
             node={\setbox\nodebox=\hbox\bgroup},
             execute at end
             node={\egroup\eqmakebox[#1][c]{\copy\nodebox}}}
            }
            
            \node 
            [unknown_op] (parent){$\bullet$}
            [child anchor=north]
                child {node [tree_op] (parallel) {$\wedge$}
                    child {node [triangle] (t1) {}}
                    child {node [tree_op] (choice) {$\circlearrowleft$}
                        child {node [triangle, selected] (frozen) {$T_i$}}
                        child {node [tree_leaf_inv] (tau) {$\tau$}}
                    }
                }
                ;
            \node [yshift=2.25em] (lca) at (t1) {$\bullet$};
            \node[left of = lca, color=gray,xshift=.6cm](){$r_c$};
            \node[above of = parent, color=gray,yshift=-.8cm](){old parent of $r_c$};
            \node[]() at (t1) {$T_{c}'$};
        \end{tikzpicture}

         \caption{Case $\{1,\infty\}$}
         \label{fig:insert-frozen-subtree-back:d}
     \end{subfigure}
    \hfill
     \begin{subfigure}[b]{0.21\textwidth}
         \centering
        \begin{tikzpicture}
            \tikzstyle{tree_leaf_inv}=[circle,draw=black,fill=black,thick,minimum size=4mm, text=white,inner sep=0pt,font=\large]

            \tikzset{
            tree_op/.style={rectangle,draw=black,fill=gray!30,thick,minimum size=4mm,inner sep=0pt},
            unknown_op/.style={circle,inner sep=-1.5pt},
            tree_subtree/.style={rectangle,draw = gray!15,fill=gray!15,thick,minimum size=4.5mm,inner sep=0pt},
            triangle/.style={inner sep=0pt,draw=gray!30, regular polygon, regular polygon sides=3, align=center,equal size=T,fill=gray!30, minimum height=4.5em},
            marking/.style={draw=black, densely dashed,thin},
            selected/.style={fill=darkgray, text=white, draw=black},
            level 1/.style={sibling distance=16mm,level distance=5mm},
            level 2/.style={sibling distance=9mm,level distance=10mm},
            level 3/.style={sibling distance=7mm},
            level 4/.style={sibling distance=11mm,level distance=11mm},
            equal size/.style={execute at begin
             node={\setbox\nodebox=\hbox\bgroup},
             execute at end
             node={\egroup\eqmakebox[#1][c]{\copy\nodebox}}}
            }
            
            \node 
            [unknown_op] (parent){$\bullet$}
            [child anchor=north]
                child {node [tree_op] (parallel) {$\wedge$}
                    child {node [triangle] (t1) {}}
                    child {node [tree_op] (choice) {$\circlearrowleft$}
                        child {node [tree_leaf_inv] (tau) {$\tau$}}
                        child {node [triangle, selected] (frozen) {$T_i$}}
                    }
                }
                ;
            \node [yshift=2.25em] (lca) at (t1) {$\bullet$};
            \node[left of = lca, color=gray,xshift=.6cm](){$r_c$};
            \node[above of = parent, color=gray,yshift=-.8cm](){old parent of $r_c$};
            \node[]() at (t1) {$T_{c}'$};
        \end{tikzpicture}

         \caption{Case $\{0,\infty\}$}
         \label{fig:insert-frozen-subtree-back:e}
     \end{subfigure}
    \caption{Four cases showing how to insert a frozen subtree $T_i$ back}
    \label{fig:insert-frozen-subtree-back}
\end{figure}
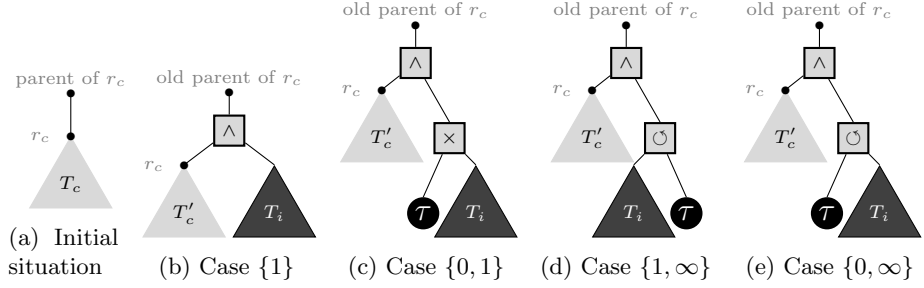

For example, reconsider \autoref{fig:running-example}.
We receive $T''{\widehat{=}}  {\rightarrow} \big({\wedge}  \big( \allowbreak {\times} \big( {\rightarrow} (a,b) ,  \allowbreak{\wedge} (c,d)    \big) ,\tau \big),\allowbreak {\rightarrow}(open^{T_2},{\circlearrowleft}(\tau,a),close^{T_2}) \big)$ (\autoref{fig:running-example-c}) after applying the IPDA (\autoref{fig:freezing_advanced}).
We observe that between opening ($open^{T_2}$) and closing ($close^{T_2}$) of the frozen subtree, the IPDA inserted a loop on $a$, i.e., ${\circlearrowleft}(\tau,a)$.
First, we calculate the lowest common ancestor (LCA) of $open^{T_2}$ and $close^{T_2}$, i.e., the sequence operator with underlying subtree ${\rightarrow}(open^{T_2},{\circlearrowleft}(\tau,a),close^{T_2})$.
Next, we do a semantical analysis of this subtree to check how often $open^{T_2}$ and $close^{T_2}$ can be replayed.
This analysis is needed since the IPDA changes the tree and $open^{T_2}$ or $close^{T_2}$ could be now skipped or executed multiple times.
In \autoref{fig:running-example-c}, $open^{T_2}$ and $close^{T_2}$ must be executed exactly once and it is neither possible to skip them nor to execute them more than once, i.e., the cardinality of $open^{T_2}$ and $close^{T_2}$ is $\{1\}$.
Hence, we apply the case $\{1\}$ visualized in \autoref{fig:insert-frozen-subtree-back:b} where $T_i$ represents the frozen subtree and $T_c'$ the determined LCA subtree after removing all nodes with label $open^{T_2}$ and $close^{T_2}$.
We obtain $T'''$ (\autoref{fig:running-example-d}) that contains the frozen subtree $T_2$ and accepts the previously added traces $\{\sigma_1{=}\langle d,c,a,b,a,e \rangle,\sigma_2{=}\langle a,b,e,a \rangle\}$ and $\sigma{=}\langle c,d,a,e,a,a \rangle$.  
Compared to the resulting tree from the baseline approach (\autoref{sec:baseline_approach}), $T'''$ is more precise because the baseline approach simply adds the frozen subtree in parallel to the resulting tree returned by the IPDA.

Subsequently, we describe the iterative reinserting of the frozen subtree(s) in general (\autoref{alg:insert-frozen-subtrees-back}).
Since we iteratively replace full executions of the frozen subtree(s) in the previously added traces $\{\sigma_1,\dots,\sigma_m\}$ (\autoref{fig:freezing_advanced}) and in the trace to be added next $\sigma$ (\autoref{fig:freezing_advanced}), we have all the intermediate projected traces available (\autoref{alg:insert-frozen-subtrees-back:line:partly-projected-1}-\ref{alg:insert-frozen-subtrees-back:line:partly-projected-2}).
First, for given replacement labels, i.e., $open^{T_i}$ and $close^{T_i}$, we calculate the LCA node $v_c$ of all leaf nodes with label $open^{T_i}$ or $close^{T_i}$ (\autoref{alg:insert-frozen-subtrees-back:line:lca}).
Next, we calculate the corresponding subtree $T_c$ with root node $r_c{=}v_c$ (\autoref{alg:insert-frozen-subtrees-back:line:lca-tree}).
The tree $T_c$ represents the first insert candidate.
Next, we semantically analyze $T_{c}$ to determine how often the nodes labeled with $open^{T_i}$ and $close^{T_i}$ have to be executed within $T_{c}$ (\autoref{alg:insert-frozen-subtrees-back:line:s-open}), i.e., Semantical Tree Analysis (STA).
Potential outcomes of the STA are: $\{1\}$ (once), $\{0,1\}$ (at most once), $\{0,\infty\}$ (zero to many), $\{1,\infty\}$ (one to many).
Next, we relabel all nodes in $T''$ that are labeled with  $open^{T_i}$ or $close^{T_i}$ by $\tau$ (\autoref{alg:insert-frozen-subtrees-back:line:tau-replacement}).
Thereby, we remove all replacement labels that correspond to the frozen subtree $T_i$ in $T''$.

Given the information $S$ from the STAs (\autoref{alg:insert-frozen-subtrees-back}, \autoref{alg:insert-frozen-subtrees-back:line:case}), we know how often the frozen subtree $T_i$ must resp. may be executed in $T_{c}$.
This allows us to define four different cases that define how the frozen subtree is inserted back, see \autoref{fig:insert-frozen-subtree-back}.
\autoref{fig:insert-frozen-subtree-back:a} shows the initial situation, i.e., we determined an insertion candidate $T_{c}$.
For example, \autoref{fig:insert-frozen-subtree-back:b} describes the case that the frozen subtree has to be executed exactly once.
After applying one of the four cases, we check if inserting the frozen subtree next to the determined subtree candidate $T_c$ is feasible (\autoref{alg:insert-frozen-subtrees-back:line:feasible}). 
If not, we undo the changes made to $T''$ (\autoref{alg:insert-frozen-subtrees-back:line:undo}) and try the next bigger subtree as insert candidate (\autoref{alg:insert-frozen-subtrees-back:line:next-candidate}).
In the worst case, we insert the frozen subtree next to the root node that is always a feasible insert candidate.

\begin{algorithm}[tb]
    
    \scriptsize
	\caption{Insert frozen subtree(s) back}
	\label{alg:insert-frozen-subtrees-back}
	\SetKwInOut{Input}{input}
	\Input{ $T''{=}(V'',E'',\lambda'',r''){\in}\mathcal{T},\allowbreak
	        T_1{=}(V_1,E_1,\lambda_1,r_1),\dots,T_n{=}(V_n,E_n,\lambda_n,r_n) {\sqsubseteq} T (n {\geq} 0),\allowbreak
	        open^{T_1},close^{T_1},\dots,open^{T_n},close^{T_n} {\in}\mathcal{R}$, $\sigma{\in}\mathcal{A}^*,\sigma_1,\dots,\sigma_m{\in}\mathcal{A}^*$
	}
	\Begin{
	    \nl \For{$T_i {\in} \{T_1,\dots,T_n\}$}{\label{alg:insert-frozen-subtrees-back:line:for}
	        
	        \nl let $\sigma^{T_1,\dots,T_i}$ be the partly projected trace $\sigma$ after replacing full executions of frozen subtrees $T_1$ to $T_i$\tcp*[r]{$\sigma^{T_1,\dots,T_n}{=}\sigma'$ (\autoref{fig:freezing_advanced})}\label{alg:insert-frozen-subtrees-back:line:partly-projected-1}
	        
	        \nl let $\{ \sigma_1^{T_1,\dots,T_i},\dots,\sigma_m^{T_1,\dots,T_i}\}$ be the partly projected, previously added traces $\sigma_1,\dots,\sigma_m$ after replacing full executions of frozen subtrees $T_1$ to $T_i$\;
	        \label{alg:insert-frozen-subtrees-back:line:partly-projected-2}

            \nl $v_c \gets$  $LCA(v_1,LCA(v_2,\dots))$ for $v_1,v_2,\dots {\in} \{ v{\in}V'' {\mid} \lambda(v){=}open^{T_i} {\lor} \lambda(v){=}close^{T_i}\}$\;\label{alg:insert-frozen-subtrees-back:line:lca}

            \nl $T_c {=} (V_c,E_c,\lambda_c,r_c) \gets {\triangle^{T''}}(v_c)$\tcp*[r]{LCA subtree is first insert candidate}
            \label{alg:insert-frozen-subtrees-back:line:lca-tree}
            
            \nl \While{ $\{ \sigma^{T_1,\dots,T_i}, \sigma_1^{T_1,\dots,T_i},\dots,\sigma_m^{T_1,\dots,T_i}\} {\nsubseteq} \mathcal{L}(T'')$ }{
            
                
                
                \nl $S \gets$ STA$(T_c, open^{T_i}) {\cap} STA(T_c, close^{T_i})$\tcp*[r]{Syntactical Tree Analysis}\label{alg:insert-frozen-subtrees-back:line:s-open}

                \nl $T_c' \gets$ relabel all nodes from $T_c$ labeled with $open^{T_i}$ or $close^{T_i}$ by $\tau$\;\label{alg:insert-frozen-subtrees-back:line:tau-replacement}
                
                \nl $T'' \gets$ apply case $S$ for frozen $T_i$ and $T_c'$\tcp*[r]{consider \autoref{fig:insert-frozen-subtree-back}}\label{alg:insert-frozen-subtrees-back:line:case}  
                
                \nl \If{
                $\{ \sigma^{T_1,\dots,T_i}, \sigma_1^{T_1,\dots,T_i},\dots,\sigma_m^{T_1,\dots,T_i}\} {\nsubseteq} \mathcal{L}(T'')$\label{alg:insert-frozen-subtrees-back:line:feasible}
                }{
                    \nl undo changes made to $T''$ (\autoref{alg:insert-frozen-subtrees-back:line:tau-replacement} and \autoref{alg:insert-frozen-subtrees-back:line:case})\;\label{alg:insert-frozen-subtrees-back:line:undo}
                    
                    \nl $T_c \gets {\triangle^{T''}}\big({p^{T''}}(r_c)\big)$\tcp*[r]{try next higher subtree as insert candidate}\label{alg:insert-frozen-subtrees-back:line:next-candidate}
                }
            }
            
	    }
	    \nl apply post-processing to $T''$\tcp*[r]{remove non-required nodes/simplify tree}
 		\nl \Return $ T''$\tcp*[r]{corresponds to $T'''$ in \autoref{fig:freezing_advanced}} 
	}
\end{algorithm}

\section{Evaluation}
\label{sec:evaluation}
\begin{figure}[t]
    \centering
     \begin{subfigure}[b]{0.49\textwidth}
        \centering
        \includegraphics[width=\textwidth,clip,trim=2.2cm 9.8cm 2.2cm 9.5cm]{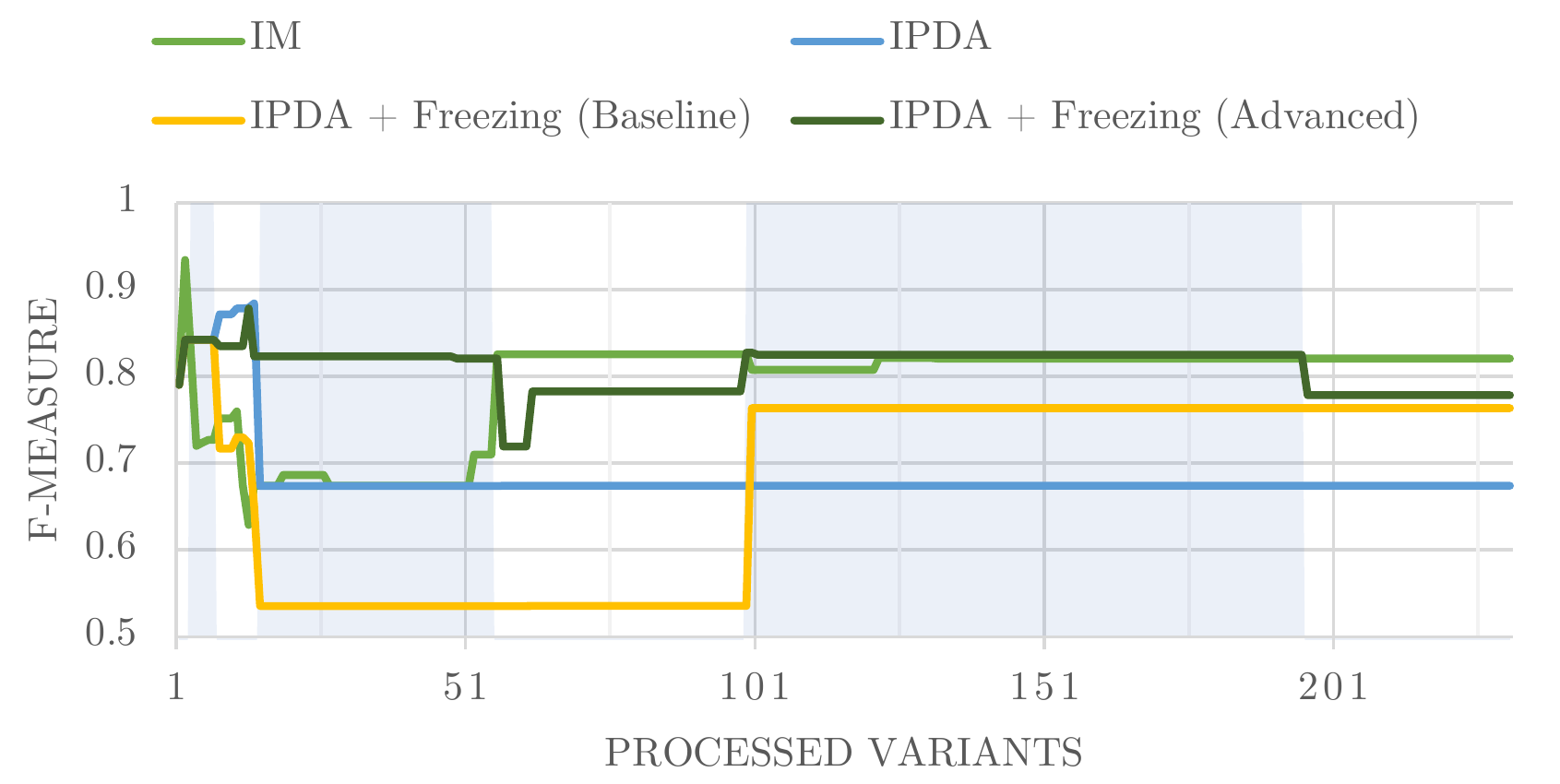}
        \caption{F-measure for experiment \texttt{rtfm\_7} \footref{footnote}}
        \label{fig:results_rtfm_a}
    \end{subfigure}
    \hfill
    \begin{subfigure}[b]{.49\textwidth}
         \centering
        \includegraphics[width=\textwidth,clip,trim=2.2cm 9.8cm 2.2cm 9.5cm]{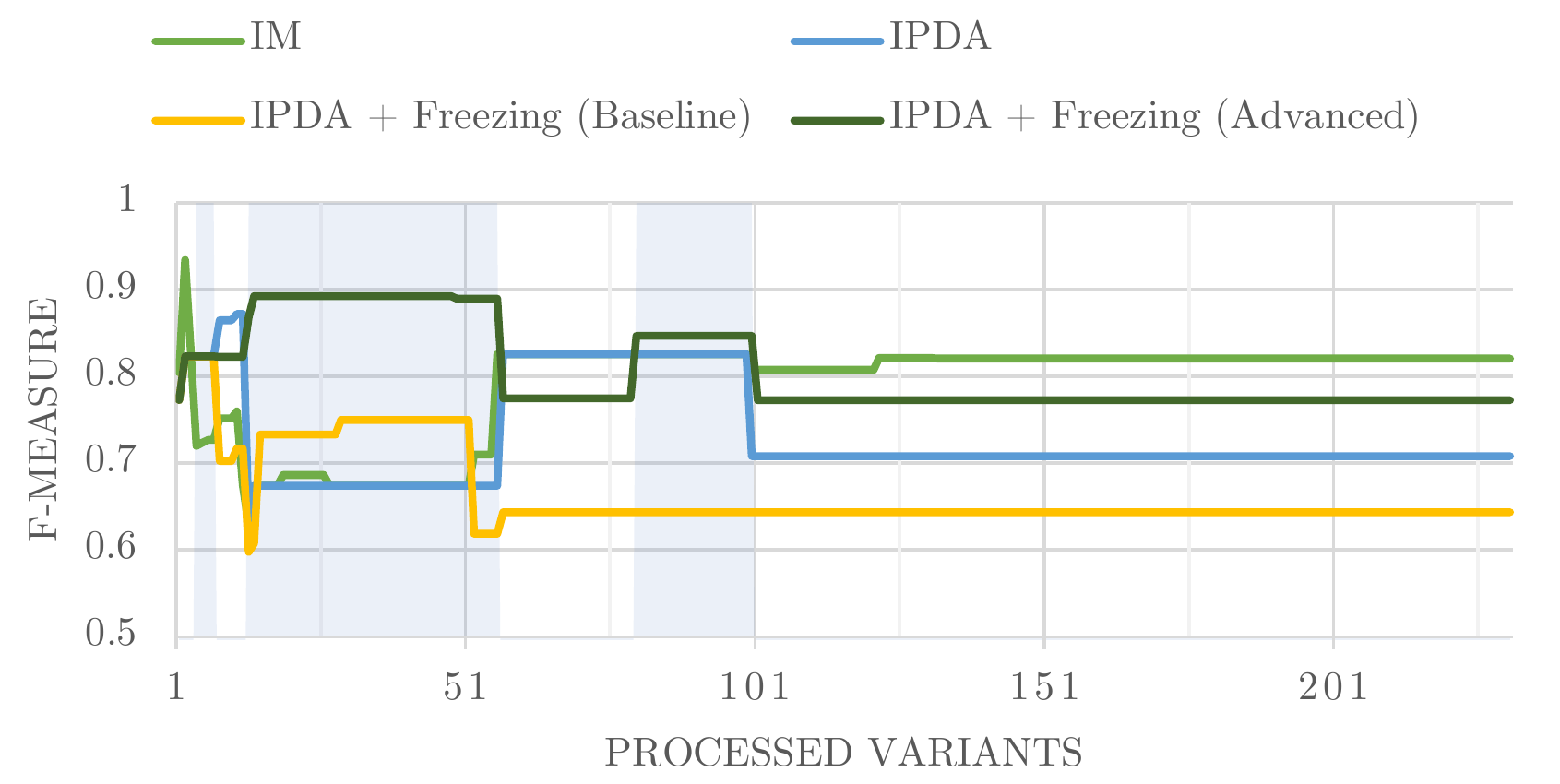}
        \caption{F-measure for experiment \texttt{rtfm\_4} \footref{footnote}}
        \label{fig:results_rtfm_b}
    \end{subfigure}

    \caption{F-measure for a real-life event log~\cite{4TU:data/road} using two different initial process models, each with a different frozen subtree. Highlighted segments indicate that the advanced freezing approach outperforms the other evaluated algorithms}
    \label{fig:results_rtfm}
\end{figure}

This section presents an experimental evaluation of the proposed freezing approach. 
We compare four different discovery approaches:
the Inductive Miner (a conventional process discovery algorithm)~\cite{DBLP:conf/apn/LeemansFA13}, an IPDA~\cite{10.1007/978-3-030-50316-1_25}, the baseline freezing approach (\autoref{sec:baseline_approach}) using the IPDA~\cite{10.1007/978-3-030-50316-1_25}, and the advanced freezing approach (\autoref{sec:advanced_approach}) using the IPDA~\cite{10.1007/978-3-030-50316-1_25}.
All four approaches have in common that they support full replay fitness, i.e., all traces given to the algorithm are accepted by the resulting tree.
We use a publicly available event log that captures a real-life process, i.e., the commonly studied \emph{Road Traffic Fine Management} (RTFM) event log~\cite{4TU:data/road}.
We sort the event log based on trace-variants frequency, i.e., most occurring trace-variant first.
For each run, i.e., \autoref{fig:results_rtfm_a} and \autoref{fig:results_rtfm_b}, we use the same initial model for all IPDA approaches and we use the same frozen subtree for both freezing approaches.
Further, we do not change the frozen subtree during incremental discovery, i.e., we freeze the same subtree in all incremental executions.
Note that in general, a user can freeze different subtree(s) after each incremental execution.
The frozen subtrees used cover a part of the reference process model presented in~\cite{DBLP:journals/computing/MannhardtLRA16}, see Figure 7, of the RTFM process.
Visualizations of the initial process trees and their frozen subtrees can be found online\footnote{\label{footnote}\url{https://github.com/fit-daniel-schuster/Freezing-Sub-Models-During-Incr-PD}}.


\autoref{fig:results_rtfm} shows the F-measure, the harmonic mean of precision and fitness, of the process trees based on the entire event log.
We observe that the advanced freezing approach clearly dominates the baseline freezing approach in both runs.
Further, we observe that the advanced freezing approach outperforms the other approaches between $3-7$, $15-55$ and $99-195$ processed trace-variants (\autoref{fig:results_rtfm_a}).
Note that in reality, \emph{incorporating all observed process behavior is often not desired} because the event data contains noise, incomplete behavior and other types of quality issues.
For instance, after integrating the first 17 most frequent trace-variants of the RTFM log, the process model covers already $99\%$ of the observed process behavior, i.e., $99\%$ of all traces are accepted by the process tree.
Comparing IPDA with the proposed advanced freezing approach (\autoref{fig:results_rtfm}), we observe that the advanced freezing approach clearly dominates IPDA in most segments.
Visualizations of all discovered process trees after each incremental execution, visualizations of the initial tree including the frozen tree, detailed statistics, and further experiments are available online\footref{footnote}.
In general, the results indicate that freezing subtrees during incremental process discovery can lead to higher quality models since we observe that the advanced freezing approach dominates the other algorithms in many segments.

\section{Conclusion}
\label{sec:conclusion}

This paper introduced a novel option to interact with a process discovery algorithm.
By allowing a user to freeze process model parts during incremental process discovery, the user gains control over the algorithm and is able to steer the algorithm.
Moreover, the proposed option to freeze parts of a process model combines conventional process discovery with data-driven process modeling and, therefore, opens a new perspective on process discovery. 
In future work, we plan to conduct research on strategies that automatically recommend process model parts which are suitable freezing candidates. 
Further, we plan to integrate the freezing approach into our incremental process discovery tool \emph{Cortado}~\cite{cortado}.

%
%
%
\bibliographystyle{splncs04}
\bibliography{main}

\end{document}